\documentclass{article}
\usepackage[utf8]{inputenc}
\usepackage[english]{babel}
\usepackage{graphicx}
\graphicspath{{images/}}

\usepackage{blindtext}
\usepackage{cvpr}
\usepackage{times}
\usepackage{epsfig}
\usepackage{graphicx}
\usepackage{amsmath}
\usepackage{amssymb}
\usepackage{scrextend}
\usepackage{enumerate}

\usepackage[utf8]{inputenc}
\usepackage[english]{babel}
\usepackage{indentfirst}
\usepackage{blindtext}
\usepackage{amssymb}
\usepackage{amsmath}
\usepackage{Notations}
\usepackage{hyperref}
\usepackage{mathtools}
\usepackage{parskip}
\usepackage{graphicx,subcaption}
\usepackage{epsfig}

\usepackage[pagebackref=true,breaklinks=true,letterpaper=true,colorlinks,bookmarks=false]{hyperref}

\usepackage{amsthm}

\usepackage{subfiles} 

\title{Subfile Example}
\author{Team Learn Overleaf}
\date{ }

\begin{document}

\section{Intro}

\title{Student-Teacher Learning from Clean Inputs to Noisy Inputs}

\author{Guanzhe Hong, Zhiyuan Mao, Xiaojun Lin, Stanley Chan\\
School of Electrical and Computer Engineering, Purdue University, West Lafayette, Indiana USA\\
{\tt\small \{hong288, mao114, linx, stanchan\}@purdue.edu}

}

\maketitle
\addtocontents{toc}{\protect\setcounter{tocdepth}{0}}
\begin{abstract}
Feature-based student-teacher learning, a training method that encourages the student's hidden features to mimic those of the teacher network, is empirically successful in transferring the knowledge from a pre-trained teacher network to the student network. Furthermore, recent empirical results demonstrate that, the teacher's features can boost the student network's generalization even when the student's input sample is corrupted by noise. However, there is a lack of theoretical insights into why and when this method of transferring knowledge can be successful between such heterogeneous tasks. We analyze this method theoretically using deep linear networks, and experimentally using nonlinear networks. We identify three vital factors to the success of the method: (1) whether the student is trained to zero training loss; (2) how knowledgeable the teacher is on the clean-input problem; (3) how the teacher decomposes its knowledge in its hidden features. Lack of proper control in any of the three factors leads to  failure of the student-teacher learning method.
\end{abstract}

\section{Introduction}
\subsection{What is student-teacher learning?}
Student-teacher learning is a form of supervised learning that uses a well-trained \emph{teacher} network to train a \emph{student} network for various low-level and high-level vision tasks. Inspired by the knowledge distillation work of Hinton et al. \cite{Hinton_knowledge_distill_2015}, Romero et al. \cite{Romero15-iclr} started a major line of experimental work demonstrating the utility of feature-based student-teacher training \cite{Heo_2019, Kim_2018, Yim_2017, Wang_2017, Srinivas_2018, Jin_2019, Tung_2019, Wang_2019, Aguilar_2019, Gnanasambandam_2020a, Chi_2020, Hong_2020, liu_2020, schwartz2021isp}.  

\begin{figure}[t]
\centering
\includegraphics[width=\linewidth]{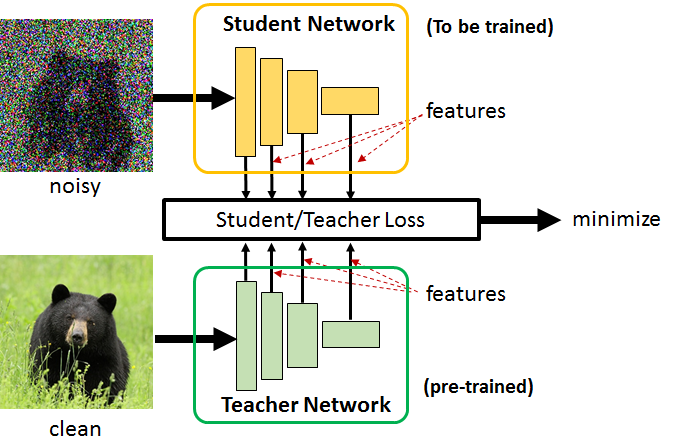}
\caption{The student-teacher training loss, computed by measuring the difference between the hidden features of the student and the teacher networks. During training, the input signal to the student network is the noisy version of the teacher's.}
\label{fig: student teacher}
\end{figure}

Figure \ref{fig: student teacher} shows an illustration of the scheme. Suppose that we want to perform classification (or regression) where the input image is corrupted by noise. In student-teacher learning, the teacher is a model trained to classify \emph{clean} images. We assume that the teacher's prediction quality is acceptable, and the features extracted by the teacher are meaningful. However, the teacher cannot handle noisy images because it has never seen one before. Student-teacher learning says that, given a pair of clean-noisy input, we can train a student by forcing the \textit{noisy} features extracted by the student to be similar to those \textit{clean} features extracted by the teacher, via a loss term known as the student-teacher loss. In some sense, the training scheme forces the student network to adjust its weights so that the features are ``denoised''. During testing, we drop the teacher and use the student for inference.

The success of student-teacher learning from clean inputs to corrupted inputs has been demonstrated in recent papers, including classification with noisy input \cite{Gnanasambandam_2020a}, low-light denoising \cite{Chi_2020}, and image dehazing \cite{Hong_2020}. However, on the theory side, there is very little analysis of why and when the hidden features of the teacher can boost the generalization power of the student. Most of the explanations in the experimental papers boil down to stating that the hidden features contain rich and abstract information about the task which the teacher solves, which could be difficult for the student network to discover on its own.

In this paper, we provide the first insights into the mechanism of feature-based student teacher learning from clean inputs to noisy inputs, for classification and regression tasks. The questions we ask are: \textit{When will student-teacher learning succeed? When will it fail? What are the contributing factors? What is the generalization capability of the student?}

The main results of our theoretical and experimental findings can be summarized in the three points below:
\boxedthm{
\begin{itemize}
    \item The student should \textbf{not} be trained to \textbf{zero training loss}.
    \item A \textbf{knowledgeable} teacher is generally preferred, but there are limitations.
    \item \textbf{Well-decomposed} knowledge leads to better knowledge transfer.
\end{itemize}
}
To verify these findings, we prove several theoretical results,  including showing how missing one or more of those can lead to failure, by studying deep linear networks. We experimentally verify these findings by studying wide nonlinear networks.

\subsection{Related works} 
Most of the existing papers on feature-based student-teacher learning are experimental in nature and include little analysis. As there already are two comprehensive review papers on these works \cite{Wang_2020a, Gou_2020}, we do not attempt to provide another one here, but instead list a few representative uses of the learning method: homogeneous-task knowledge transfer techniques, which include general-purpose model compression \cite{Heo_2019, Kim_2018, Yim_2017, Jin_2019, Tung_2019, Aguilar_2019}, compression of object detection models \cite{Wang_2019}, performance improvement on small datasets \cite{Yim_2017}; heterogeneous-task knowledge transfer techniques similar to that depicted in Figure \ref{fig: student teacher}: the student's input is usually corrupted by noise \cite{Gnanasambandam_2020a, Srinivas_2018}, blur \cite{Hong_2020}, noise with motion \cite{Chi_2020}, etc.

On the theory front, there are three papers that are most related to our work. The first is \cite{Vapnik_2015}, which formulates student-teacher learning in the framework of ``privileged information''. We found two limitations of the work: first, it only focuses on student-teacher learning using kernel classifiers and not neural networks; second, it does not clearly identify and elaborate on the factors that lead to the success and failure cases of student-teacher learning. The second paper of interest is \cite{Rahbar_2020}, which focuses on the \textit{training dynamics} of student-teacher learning, while our work focuses on the \textit{generalization} performance of the trained student. The third one is \cite{Phuong_2019}. Aside from studying \textit{target}-based (instead of feature-based) student-teacher learning for deep linear networks, there are two additional differences between their work and ours: they only focus on the case that the teacher and student's tasks are identical, while we assume the student faces noisy inputs, moreover, some of their messages appear opposite to ours, e.g. from their results, early stopping the student network is not necessary, and might, in fact, harm the student's generalization performance, while we claim the opposite.

\subsection{Scope and limitations}
We acknowledge that, due to the varieties and use cases of student-teacher learning, a single paper cannot analyze them all. In this paper, we focus on the case depicted in Figure \ref{fig: student teacher}: the teacher and student network have identical architecture, no transform is applied to their features, and they solve the same type of task except that the student's input is the noisy version of the teacher's. We do \textit{not} study the learning method in other situations, such as model compression. Moreover, our focus is on the generalization performance of the student, not its training dynamics.

\section{Background}

We first introduce the notations that shall be used throughout this paper. We denote the clean training samples $\{(\vx_i, \vy_i)\}_{i=1}^{N_s}\subset\mathbb{R}^{d_x} \times \mathbb{R}^{d_y}$, and the noise vectors $\{\vepsilon_i\}_{i=1}^{N_s}\subset\mathbb{R}^{d_x}$. For matrix $\mM$, we use $[\mM]_{i,j}$ to denote the $(i,j)$ entry of $\mM$, and $[\mM]_{i,:}$ and $[\mM]_{:,j}$ to denote the $i$-th row and $j$-th column of $\mM$. For convenience, we define matrices $[\mX]_{:,i} = \vx_i$, $[\mX_{\epsilon}]_{:,i} = \vx_i+\vepsilon_i$, and $[\mY]_{:,i} = \vy_i$.

We write an $L$-layer neural network $\vf(\mW_1, ..., \mW_L; \cdot):\mathbb{R}^{d_x}\to\mathbb{R}^{d_y}$ as (for simplicity we skip the bias terms):
\begin{equation*}
    \vf(\mW_1, ..., \mW_L; \vx) = \sigma(\mW_L\sigma(\mW_{L-1}...\sigma(\mW_1\vx)...)
\end{equation*}
where the $\mW_i$'s are the weights, and $\sigma(\cdot)$ is the activation function. In this model, if $\sigma(\cdot)$ is the identity function, we have a deep linear network as a special case. Deep linear networks have been used in the theoretical literature of neural networks \cite{Saxe_2014, Arora_2018a, Arora_2018b, Arora_2019, Kawaguchi_2016, Phuong_2019}, as they are often more analytically tractable than their nonlinear counterparts, and help provide insights on the mechanisms of the nonlinear networks. We denote $\mW_{\textbf{L}} = \prod_{i=1}^L\mW_i$.

While we will demonstrate numerical results for $L$-layer linear (and nonlinear) networks, to make our theoretical analysis tractable, we make the following assumptions:
\boxedthm{
Assumptions for theoretical results in this paper:
\begin{itemize}
\item Assumption 1: The student and the teacher share the \textit{same} 2-layer architecture: shallow ($L = 2$), fully-connected, and the dimension of the single hidden layer is $m$.
\item Assumption 2: Noise is only applied to the \textit{input} of the student, the targets are always noiseless.
\end{itemize}
}

In terms of the training losses of the student network, we denote $\widehat{\calL}_{\text{base}}(\mW_1, \mW_2)$ as the \textit{base training loss}:
\begin{equation}
    \widehat{\calL}_{\text{base}}(\mW_1, \mW_2) = \sum_{i=1}^{N_s} \ell(\vf(\mW_1, \mW_2; \underset{\textcolor{blue}{\text{noisy input}}}{\underbrace{\vx_i + \vepsilon_i}}),  \underset{\text{\textcolor{blue}{clean label}}}{\underbrace{\vy_i}})
\label{eq: base training loss}
\end{equation}
where $\ell:\mathbb{R}^{d_y}\times\mathbb{R}^{d_y} \to \mathbb{R}_{\ge 0}$ can be, for instance, the square loss. Moreover, we define the \textit{student-teacher (ST) training loss} as follows:
\begin{equation}
\begin{aligned}
    &\widehat{\calL}_{\text{st}}(\mW_1, \mW_2)
    = 
    \underset{\textcolor{blue}{\text{base training loss}}}{\underbrace{\widehat{\calL}_{\text{base}}(\mW_1, \mW_2)}} + \\
    &\qquad\qquad \lambda
    \underset{\textcolor{blue}{\text{feature difference loss}}}{\underbrace{\sum_{i=1}^{N_s}\|\sigma(\mW_1(\vx_i+\vepsilon_i)) - \sigma(\widetilde{\mW}_1\vx_i)\|_2^2}} 
\label{eq: ST loss}
\end{aligned}
\end{equation}
where the $\widetilde{\mW}_i$'s are the weights of the teacher network. The feature difference loss for 2-layer networks can be easily generalized to deeper networks: for every $h \in \{1, ..., L-1\}$, sum the $\ell_2$ difference between the hidden features of layer $h$ from the student and teacher networks.

During testing, we evaluate the student network's generalization performance using the \textit{base testing loss}:
\begin{equation}
    \calL_{\text{test}}(\mW_1, \mW_2) \bydef \E_{\vx,\vy,\vepsilon}\left[\ell(\vf(\mW_1, \mW_2; \vx + \vepsilon), \vy)\right].
\end{equation}
Unlike many existing experimental works, we do not apply any additional operation to the hidden features of the student and teacher networks. We choose the particular student-teacher loss because we wish to study this training method in its simplest form. Furthermore, this form of student-teacher loss is close to the ones used in \cite{Gnanasambandam_2020a, Chi_2020}.

\section{Message I: Do not train student to zero loss}
The student-teacher loss \eqref{eq: ST loss} can be viewed as the base loss \eqref{eq: base training loss} regularized by the feature difference loss \cite{Romero15-iclr, Gnanasambandam_2020a}. A natural question then arises: since we are already regularizing the base loss, shall we train the overall student-teacher loss to zero so that we have the optimal student-teacher solution? The answer is \textit{no}. The main results are stated as follows.

\boxedthm{
\textbf{Message I}: Do not train the student to zero training loss. 
\begin{itemize}
    \item Section 3.1: If the deep linear network is over-parametrized $N_s < d_x$, training the student  until zero training loss using \eqref{eq: ST loss} will return a solution close to the base one (Theorem 1). Similar conclusion holds for $N_s \ge d_x$ (Theorem 2). 
    \item Section 3.2: An early-stopped student trained with \eqref{eq: ST loss} has better test error than one trained to convergence.
\end{itemize}
}

\subsection{Theoretical insights from linear networks}
To prove the theoretical results in this sub-section, we assume that $\sigma(\cdot)$ is the identity function, and the base training and testing loss are the MSE loss, i.e. $\ell(\widehat{\vy}, \vy) = \|\widehat{\vy} - \vy\|_2^2$. We explicitly characterize how close the solutions of the MSE and S/T losses are.

\begin{theorem}
\label{thm: student reg, undersampling}
Let $L = 2$. Suppose the student's sample amount $N_s < d_x$, $\{\vx\}_{i=1}^{N_s}$ and $\{\vepsilon\}_{i=1}^{N_s}$ are sampled independently from continuous distributions, and the optimizer is gradient flow. Denote $\mW^{\text{base}}_i(t)$ and $\mW^{\text{st}}_{i}(t)$ as the weights for the student network trained with the base loss \eqref{eq: base training loss} and the student-teacher loss \eqref{eq: ST loss}, respectively. 

Assume that the following statements are true:
\begin{enumerate}[i]
    \item There exists some $\delta > 0$ such that $\|\mW_i^{\text{base}}(0)\|_F \le \delta$ and $\|\mW_i^{\text{st}}(0)\|_F \le \delta$ for all $i$;
    \item The teacher network minimizes the training loss for clean data $\sum_{i=1}^{N_s} \ell(\vf(\widetilde{\mW}_1, \widetilde{\mW_2}); \vx_i)$;
    \item Gradient flow successfully converges to a global minimizer for both the MSE- and ST-trained networks
\end{enumerate}
With mild assumptions on the initialized weights and the gradient flow dynamics induced by the two losses, and with $\delta$ sufficiently small, the following is true almost surely:
\begin{equation}
    \lim_{t\to\infty} \|\mW_{\textbf{L}}^{\text{base}}(t) - \mW_{\textbf{L}}^{\text{st}}(t)\|_F \le C\delta
\end{equation}
for some constant $C$ that is independent of $\delta$.
\end{theorem}
\emph{Proof}. See supplementary materials.

The implication of the theorem is the following. When we initialize the student's weights with small norms, which is a standard practice \cite{Bengio_2010, He_2015}, and if the teacher satisfies several mild assumptions, then the final solution reached by the MSE- and the student-teacher-induced gradient flow are very close to each other. In other words, \textit{using student-teacher training does not help if we train to zero loss}. 

We elaborate on some of the assumptions. The assumption $N_s < d_x$ causes the optimization problem to be underdetermined, leading to nonunique global minima to the base and student-teacher problems. Thus, we need to consider solutions that the gradient flow optimizer chooses. Assumption (iii) simplifies our analysis and is similar to the one made in \cite{Arora_2019}. It helps us to focus on the end result of the training rather than the dynamics.

We observe similar phenomenon when $N_s \ge d_x$, albeit with stricter assumptions on the two networks.

\begin{theorem}
\label{thm: student reg, oversampling}
Suppose $N_s \ge d_x$. Assume that $L=2$,  $\text{span}\left(\{\vx_i+\vepsilon_i\}_{i=1}^{N_s}\right) = \mathbb{R}^{d_x}$, the teacher network can perfectly interpolate the clean training samples, and the dimension of the hidden space $m$ is no less than $\text{rank}(\mY\mX_{\vepsilon}^T(\mX_{\vepsilon}\mX_{\vepsilon}^T)^{-1})$. Then the global minimizers of MSE and S/T satisfy:
\begin{equation}
    \mW_{\textbf{L}}^{\text{base}} = \mW_{\textbf{L}}^{\text{st}} = \mY\mX_{\vepsilon}^T(\mX_{\vepsilon}\mX_{\vepsilon}^T)^{-1}
\end{equation}
\end{theorem}
\emph{Proof}. See supplementary materials.

Theorem \ref{thm: student reg, oversampling} tells us that when the teacher network has zero training error on the clean-input task, and the student possesses sufficient capacity, MSE and S/T learning produce exactly the same student network. Additionally, as proven in the supplementary materials, very similar versions of the current and previous theorem hold even if the teacher's activation function is \textit{not} the identity. It can be any function. 

The two theorems show that, even though the feature difference loss in \eqref{eq: ST loss} can be viewed as a regularizer, it is important to add other regularizers or use early stopping so that \eqref{eq: ST loss} can provide benefit to the student.

\subsection{Experimental evidence}

Since the theoretical analysis has provided justifications to the linear networks, in this sub-section, we conduct a numerical experiment on nonlinear networks to strengthen our claims.  

\textbf{Choices of teacher and student}. We consider a teacher and a student that both are shallow and wide fully-connected ReLU networks with hidden dimension $m = 20,000$, input dimension $d_x = 500$, and output dimension $d_y = 1$. We assume that the teacher network is the ground truth here, and the teacher's layers are set by the Xavier Normal initialization in PyTorch, i.e. each entry of $\widetilde{\mW}_1$ is sampled from $\calN(0, 2/(d_x + m))$, and each entry of $\widetilde{\mW}_2$ is sampled from $\calN(0, 2/(d_y + m))$. The clean input data $\vx\sim\calN(\vzero, \mI)$, and the noise $\vepsilon\sim\calN(\vzero, \sigma_{\epsilon}^2\mI)$, with $\sigma_{\epsilon}=0.5$. The loss $\ell(\cdot, \cdot)$ is the square loss, so the learning task is MSE regression. All networks are optimized with batch gradient descent.

\textbf{Experimental setting}. The goal of the experiment is to demonstrate the benefit of early stopping to the trained student's testing error. We first randomly sample $\{\vx_i+\vepsilon_i\}_{i=1}^{N_s}$ and compute the $\{y_i\}_{i=1}^{N_s}$. To train the student network using \eqref{eq: ST loss}, we carry out parameter sweep over $\lambda$ in \eqref{eq: ST loss}, and for each $\lambda$ used, we record that student's best testing error during training and at the end of training. Note that all of these trained students use Xavier normal initialization with the same random seed and the same training samples. We found that the best test error always occurs \textit{during} training, i.e. early stopping is necessary. Out of all the early-stopped networks trained with different $\lambda$'s, we pick out the one that has the best early-stopped test error, and plot this error on the ``Early-Stopped'' curve, and that network's error at the end of training on the ``Zero Training Loss'' curve. Finally, for comparison purposes, for all the $N_s$'s we choose, we also train student networks using the base loss \eqref{eq: base training loss}, with the same samples and initialization, and early-stopped for optimal generalization.

\begin{figure}[t]
  \centering
    \includegraphics[width=1.0\linewidth]{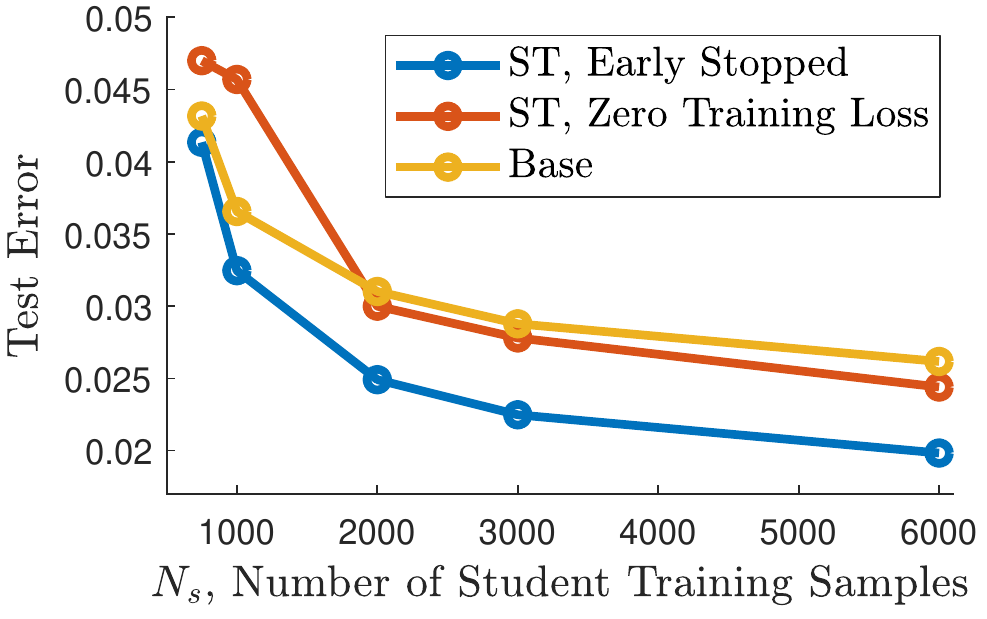}
  \caption{Testing error of student networks trained with the student-teacher loss \eqref{eq: ST loss}, with and without early stopping, and student network trained with the base loss \eqref{eq: base training loss}. The figure shows that early stopping is necessary for student-teacher learning to have significant improvement over baseline learning. }
  \label{fig: 3.2}
\end{figure}

\textbf{Conclusion}. The experimental results are depicted in Figure \ref{fig: 3.2}. The horizontal axis is $N_s$, i.e. the amount of noisy training samples available to the student, and the vertical axis is the test error of the trained student. Indeed, the early-stopped students trained with \eqref{eq: ST loss} can outperform both the ``zero-training-loss'' student and the baseline student, which supports the necessity of early-stopping the student.

\section{Message II: Use a knowledgeable teacher}

In this section, we shift our attention to the teacher. We consider the following questions: How knowledgeable should the teacher be  (1) if we want student-teacher learning to generalize better than using the base learning? (2) if the input data becomes noisier so that more help from the teacher is needed? To quantify the level of a teacher's ``knowledge'', we use the number of training samples seen by the teacher as a proxy. The intuition is that if the teacher sees more (clean) samples, it should be more knowledgeable.
\boxedthm{
\textbf{Message II}: For any teacher pre-trained with a finite amount of data, there exists an operating regime for the student-teacher learning to be effective. The regime depends on the number of training samples available to the teacher and student. Generally, a more knowledgeable teacher is preferred. 
\begin{itemize}
    \item Section 4.2: If more training samples are available to the students, the teacher needs to be more knowledgeable for the student-teacher learning to be effective. 
\end{itemize}

}

\boxedthm{
\begin{itemize}
    \item Section 4.3: If the student's task becomes more difficult (i.e. the noise level is higher), the teacher needs to be more knowledgeable in order to help the student. 
\end{itemize}

}

\subsection{Experimental setting}
We conduct several experiments using deep nonlinear convolutional networks to verify the message. Before we dive into the details, we define a few notations, as shown in Table~\ref{table: notations for Section 4}.

\begin{table}[h]
\centering
\begin{tabular}{ll}
\hline\hline
$N_s$ & number of noisy samples/class for student \\
$N_t$ & number of clean samples/class for teacher\\
$\vf_{\text{t}}(N_t)$ & teacher trained with $N_t$ clean samples\\
$\vf_{\text{st}}(N_t, N_s)$ & student trained with $N_s$ noisy samples \\
& and $\vf_t(N_t)$ using student-teacher loss \eqref{eq: ST loss}\\
$\vf_{\text{base}}(N_s)$ & same as $\vf_{\text{st}}$ but trained using base loss \eqref{eq: base training loss} \\
$E_\text{t}(N_t)$ & testing error for $\vf_{\text{t}}(N_t)$\\
$E_\text{st}(N_t, N_s)$ & testing error for $\vf_{\text{st}}(N_t, N_s)$\\
$E_\text{base}(N_s)$ & testing error for $\vf_{\text{base}}(N_s)$\\
\hline
\end{tabular}
\caption{Notations for Section 4.}
\label{table: notations for Section 4}
\end{table}

The goal of this experiment is to show the regime where student-teacher learning is beneficial. To this end, we aim to visualize the equation
\begin{equation}
    E_{\text{st}}(N_t,N_s) \le (1-\delta)E_{\text{base}}(N_s),
\end{equation}
for some hyper-parameter $\delta > 0$. Given noise level $\sigma_{\epsilon}$, this equation depends on how knowledgeable the teachers is (based on $N_t$), and how many samples the student can see ($N_s$). 

For this experiment, we consider a classification problem on CIFAR10 dataset. We use ResNet-18 as the backbone for both student and teacher networks. The feature-difference loss is applied to the output of each stage of the network, and we fix the hyper-parameter $\lambda$ in \eqref{eq: ST loss} to 0.001 for all training instances, as it already yields good testing error. Optimization-wise, both the student and the teacher networks are trained with SGD optimizer from scratch for 300 epochs, and the learning rate is set to 0.01 initially and is divided by 10 after every 100 epochs. To make early stopping possible, we allocate 2000 images from the testing set to form a validation set. The best model on the validation set from the 300 epochs is saved. 

To minimize the random effect during the training process, we do not use any dropout or data augmentation. We also make sure that the networks with the same training sample amount ($N_s$ or $N_t$) are trained with the \textit{same} subset of images. Each model is trained 5 times with different random seeds, and the average performance is reported.

\subsection{Operating regime of student-teacher learning}

\begin{figure}[h!]
\centering\includegraphics[width=1.0\linewidth]{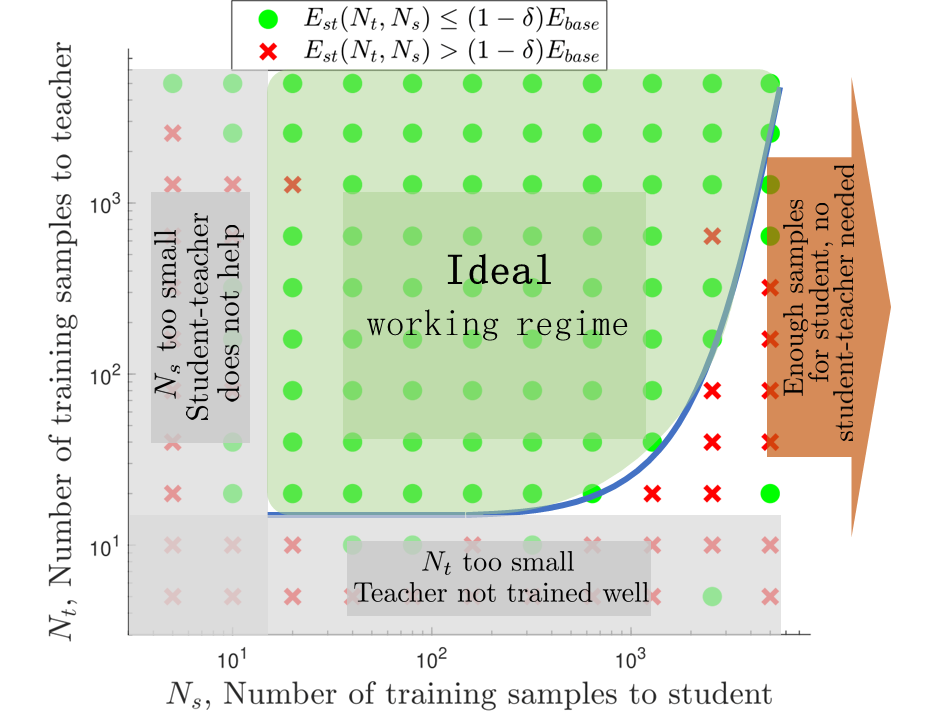}
\caption{Operating regime of student-teacher learning. Green circles \textcolor{green}{$\bullet$} represent the actual numerical experiment where $E_{\text{st}} \le (1-\delta)E_{\text{base}}$, and red crosses \textcolor{red}{$\times$} represent $E_{\text{st}} > (1-\delta)E_{\text{base}}$. We highlight regions where students-teacher learning can be benefited. If $N_t$ is too small, $N_s$ is too small, or $N_s$ is too large, there is little benefit of student-teacher learning.}
 \label{fig: 4.2}
\end{figure}

Understanding the operating regime can be broken down into two sub-questions:
\begin{enumerate}
\setlength\itemsep{0ex}
\item[(1a)] Is there a range of $N_s$ such that regardless of how big $N_t$ is, student-teacher learning simply cannot beat base learning? 
\item[(1b)] Away from the regime in (1a), as $N_s$ varies, how should $N_t$, the teacher's training sample quantity, change such that student-teacher learning can outperform base learning? 
\end{enumerate}

\textbf{Generation of Figure \ref{fig: 4.2}.} The answers to the above questions can be obtained from Figure \ref{fig: 4.2}. The figure's x-axis is $N_s$ and y-axis is $N_t$. Parameter-wise, the data in the figure is generated by varying $N_s$ and $N_t$, while keeping $\sigma_{\epsilon}$ fixed to 0.5. Procedure-wise, we first select two sets, $\calN_t$ and $\calN_s\subset\mathbb{N}$. For every $N_t\in\calN_t$, we train a teacher network $\vf_{\text{t}}(N_t)$, early-stopped to have the best testing error on the clean-input task. Then for each $N_s \in \calN_s$ and each $\vf_{\text{t}}(N_t)$, we train a student network $\vf_{\text{st}}(N_t, N_s)$ using the student-teacher loss \eqref{eq: ST loss}, and train a $\vf_{\text{base}}(N_s)$ with the base loss \eqref{eq: base training loss}. The above experiment is repeated over different $N_t$'s. Now, we fix $\delta = 0.02$, and compare $E_{\text{st}}(N_t, N_s)$ against $E_{\text{base}}(N_s)$ over all the pairs of $N_t$ and $N_s$. If $E_{\text{st}}(N_t, N_s) \le (1-\delta)E_{\text{base}}(N_s)$, we mark the position $(N_t, N_s)$ with a green dot in the figure, otherwise, we mark it with a red cross. For clearer visualization, we use color blocks to emphasize the important regions in the figure.

\textbf{Answering question (1a)}. In Figure \ref{fig: 4.2}, we see that when $N_s$ is too small, the region is filled with red crosses, i.e. student-teacher learning cannot outperform baseline learning regardless of what $N_t$ is. Intuitively speaking, when $N_s$ is too small, it simply is impossible for the student to extract any meaningful pattern from its training data, regardless of how good the teacher's features are.

\textbf{Answering question (1b)}. Figure \ref{fig: 4.2} shows that, as $N_s$ increases, the lower boundary of the green region keeps moving upward, which means that $N_t$  must also increase for student-teacher learning to beat the baseline. This phenomenon is also intuitive to understand: as the student sees more and more training samples, its ability to capture the target-relevant information in the noisy input would also grow, so it should also have higher demand on how much target-relevant hidden-feature information the teacher provides about the clean input.

\subsection{The influence of student's task difficulty}
Another related question is the following: 
\begin{enumerate}
    \item[(2)] How knowledgeable should the teacher be when the student needs to handle a difficult task, so that student-teacher learning is effective?
\end{enumerate}
To answer the above question, we conduct the following experiment. We fix $N_s = 320$ and $\delta=0.04$, increase $\sigma_{\epsilon}$ from $0.1$ to $0.5$ by steps of $0.1$, and observe how $N_t$ needs to change in order for $E_{\text{st}}(N_t, N_s)\le (1-\delta)E_{\text{base}}(N_s)$ to be maintained. The result is shown in Table \ref{table: difficulty}. Note that the $N_t$'s vary by steps of $100$.

\textbf{Interpreting Table \ref{table: difficulty}}. It can be seen that as $\sigma_{\epsilon}$ increases, $N_t$ must also increase in order for $E_{\text{st}}(N_t, N_s)\le (1-\delta)E_{\text{base}}(N_s)$. Intuitively speaking, as the noise in the student's training input samples becomes heavier, it becomes harder for the student to extract target-relevant patterns from the input, as the noise obscures the clean patterns. This in turn means that the teacher needs to give the student information of greater clarity in order to help the student, and this boils down to an increase in $N_t$.

\begin{table}[h]
\centering
\begin{tabular}{c c c c c c}
\hline\hline
$\sigma_\epsilon$ & 0.1 & 0.2 & 0.3 & 0.4 & 0.5\\
$N_t$ & 200 & 300 & 500 & 700 & 800\\

\hline\\
\end{tabular}
\caption{Minimum training samples $N_t$ required at each $\sigma_\epsilon$ level. }
\label{table: difficulty}
\end{table}

\subsection{Summary} 
The experimental results above suggest a few important observations. Firstly, a large $N_t$ (i.e., a more knowledgeable teacher) is generally beneficial. Secondly, if $N_s$ is too small or too large, the student-teacher offers little benefit. Thirdly, a larger $\sigma_{\epsilon}$ generally demands a more knowledgeable teacher.

\section{Message III: Well-decomposed knowledge leads to better knowledge transfer}
In Section 4, we observed that when $N_t$ is large, student-teacher learning usually outperforms the baseline learning. However, the following question remains unanswered: \textit{Does a good teacher only mean someone with a low testing error?} Intuitively we would think yes, because low-test-error means that the teacher performs well on its own task. However, having a low testing error does not mean that the teacher can \emph{teach}. In fact, student-teacher learning benefits from a ``divide-and-conquer'' strategy. If the knowledge can be decomposed into smaller pieces, student-teacher learning tends to perform better.

\boxedthm{
\textbf{Message III}: Student-teacher learning improves if the teacher can decompose knowledge into smaller pieces. 
\begin{itemize}
    \item Section 5.2: If the teacher's hidden features have sufficiently low complexity, then it is easy for the student to mimic the teacher's features, hence resulting in low test error on the noisy task (Theorem 3);
    \item Section 5.3: When $N_s$ is not too small, a similar phenomenon happens for nonlinear networks.
\end{itemize}
}

\subsection{Theoretical setting}
We first need to settle on a way to quantify how decomposed the knowledge is. Since the concept of ``knowledge'' itself is vague, we acknowledge that any definition surrounding its decomposition would have some degree of arbitrariness. 

\textbf{Unit of knowledge --- how neurons are grouped.} We adopt the following definition of \textit{units of knowledge} in the hidden layer. For linear networks, the unit is any hidden neuron with weight that has sparsity level of 1, i.e. only one of its entries is nonzero. This choice fits the intuition of the simplest linear transform possible, and is compatible with the popular LASSO regression model. We shall further elaborate on this in section 5.2.

For ReLU networks, we treat any hidden ReLU neuron as one unit of knowledge. When outputs from more ReLU neurons are linearly combined together, we treat them as larger units of knowledge as they form more complex piecewise linear functions. This observation is further supported on wide fully-connected ReLU networks. If such a network was trained with gradient descent and initialized with standard schemes, such as the Xavier Normal initialization, the hidden neurons' weights would be close to their random initialization \cite{Arora_2019b, Jacot_2018}. Therefore, given a group of these neurons, as long as the group is not too large, their weights are unlikely to be col-linear, so linearly combining the outputs of them indeed create more complex functions.

\textbf{Additional assumptions}. To provide a concrete theoretical result, we make several additional assumptions:

\begin{enumerate}[i]
    \item We assume that the teacher network has \textbf{zero test error}. This is the best-case-scenario in Section 4. 
    \item We focus on the \textit{simplified student-teacher training loss}, defined as follows: 
\begin{equation}
\begin{aligned}
    &\widehat{\calL}_{\text{st}}^{\text{simp}}(\mW_1) \\
    &\quad = \sum_{i=1}^{N_s} \Big\|\mP\big[\sigma(\mW_1(\vx_i+\vepsilon_i)) - \sigma(\widetilde{\mW}_1\vx_i)\big]\Big\|_2^2
\end{aligned}
\label{eq: simplified ST loss}
\end{equation}

\begin{figure}[t!]
  \centering
  \begin{subfigure}
    \centering\includegraphics[width=1.0\linewidth]{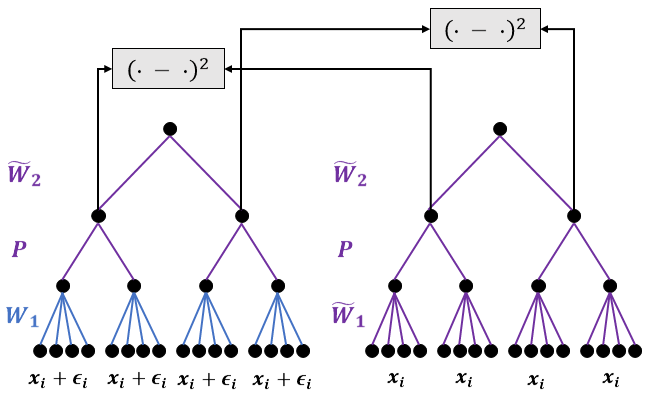}
  \end{subfigure}
  \caption{An illustration of the simplified student-teacher loss \eqref{eq: simplified ST loss}. Here, $d_x = 4$, $m=4$, and $g = 2$. Notice that the square difference is taken between the pooled features of the student and teacher networks.}
  \label{fig: section 5 scheme}
\end{figure}

An illustration of the above loss is shown in Figure \ref{fig: section 5 scheme}. The base loss $\widehat{\calL}_{\text{base}}(\mW_1, \mW_2)$, which provides target information, is not present here. The matrix $\mP\in\mathbb{R}^{(m/g) \times m}$, where $g\in\mathbb{N}$ is a divisor of $m$, and $\mP_{i,j} = 1$ if $j\in\{i g, ..., (i+1) g\}$, and zero everywhere else. Multiplication with $\mP$ essentially \textit{sums every $g$ neurons' output}, similar to how average pooling works in convolutional neural networks. We treat $g$ as a proxy of how decomposed the teacher's features are: the larger it is, the less decomposed the features are. 
\item We fix $\mW_2 = \widetilde{\mW}_2$, i.e. the second layer of the student is fixed to be identical to the teacher's, and only $\mW_1$ is trainable. At inference, the student computes $\widetilde{\mW}_2\mP\sigma(\mW_1(\vx+\vepsilon))$, and teacher computes $\widetilde{\mW}_2\mP\sigma(\widetilde{\mW}_1\vx)$.
\item We assume that the entries in the noise vectors $\vepsilon$ are all zero-mean Gaussian random variables with variance $\sigma_{\epsilon}^2$.
\end{enumerate}

\subsection{Theoretical analysis via LASSO}
We formulate the knowledge decomposition analysis via LASSO, because it offers the most natural (and clean) analytical results. We use the identity for the activation function $\sigma(\cdot)$. For simplicity, we use $d_y = 1$, and use the square loss for $\ell(\cdot, \cdot)$. Thus, our learning problem reduces to linear regression. Following the suggestion of Section 3, we impose an $\ell_1$-regularization onto the student so that it becomes a LASSO.

\begin{theorem} 
\label{theorem: S/T individual lasso error}
Assume assumptions (i)-(iv) in Section 5.1, and consider the following conditions:
\begin{itemize}
\setlength\itemsep{0ex}
    \item The ground truth is a \textit{linear} model characterized by the vector $\vbeta^*\in\mathbb{R}^{d_x}$, and without loss of generality, only the first $s$ entries are nonzero.
    \item The hidden dimension of the networks $m$ is equal to the number of non-zeros $s$.
    \item The weights of the teacher satisfy $[\widetilde{\mW}_2]_i = 1$ for all $i=1, ..., s/g$;  $[\widetilde{\mW}_1]_{i,i} = \beta^*_i$ for $i=1, ..., s$, and the remaining entries are all zeros. Essentially, the $s/g$ groups of pooled teacher neurons in \eqref{eq: simplified ST loss} each has $g$ distinct entries from $\vbeta^*$. 
    \item The number of samples satisfies \footnote{We hide constants coming from the technical LASSO analysis with  $\widetilde{}$  on top of $\calO$ and $\Omega$. \label{footnote: tilde notation}} $N_s \in \widetilde{\Omega} \left(g^2 \log(d_x)\right)$. 
    \item The samples $\{\vx\}_{i=1}^{N_s}$ and some of the parameters above satisfy certain technical conditions (for LASSO analysis).
\end{itemize}
Then, with high probability, the student network which minimizes \eqref{eq: simplified ST loss} achieves mean square test error
\begin{align}
    & \E\left[\left(\widetilde{\mW}_2\mP\mW_1(\vx+\vepsilon) -\vbeta^{*T}\vx \right)^2\right]  = 
    \widetilde{\calO}\footref{footnote: tilde notation}\bigg(\frac{\sigma_{\epsilon}^2\|\vbeta^*\|_2^2}{1+\sigma_{\epsilon}^2} \bigg).
\end{align}
\end{theorem}
\emph{Proof}. See supplementary materials.

\begin{figure}[t!]
  \centering
  \begin{subfigure}
    \centering\includegraphics[width=1.0\linewidth]{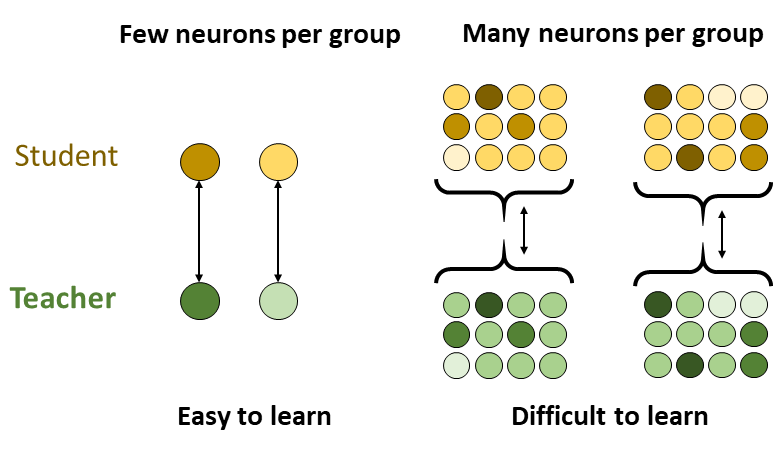}
  \end{subfigure}
  \caption{Small-$g$ (left) vs. large-$g$ (right) student-teacher learning.}
  \label{fig: decomp illustration}
\end{figure}

\textbf{Interpreting the theorem}. Note that, when $g$ is small, the above error can be quite close to the optimal test error $\sigma_{\epsilon}^2\|\vbeta^*\|_F^2/(1+\sigma_{\epsilon}^2)$, shown in the supplementary notes. More importantly, the required sample amount $N_s$ is independent of $s$, the ``complexity'' of the ground truth linear model. In contrast, if we only use the targets to train the student, standard LASSO literature suggests that $N_s$ should at least be $\Omega(s)$ to achieve nontrivial generalization \cite{Meinshausen_2006, Hastie_2015, Buhlmann_2011}. Thus, by decomposing the teacher's knowledge into simple units, student-teacher learning can succeed with much fewer samples than base learning. See the supplementary notes for experimental demonstrations of students trained by \eqref{eq: simplified ST loss} outperforming those trained with targets by a significant margin. 

Besides the fact that the teacher has zero testing error, the key reason behind this effective learning is the ``divide-and-conquer'' strategy adopted by \eqref{eq: simplified ST loss}. This idea is roughly illustrated in Figure \ref{fig: decomp illustration}. Imagine that each small disk represents a hidden neuron of a network, and the left and right sides represent two ways of teaching the student. The left is essentially giving the student neurons simple pieces of information \textit{one at a time}, while the right floods the student neurons complex information pooled from many teacher neurons \textit{all at once}. The left side clearly represents a better way of teaching, and corresponds to a choice of small $g$. 

Now, let us consider the more precise example in Figure \ref{fig: section 5 scheme}, in which $d_x = 4$, $m=4$, $g=2$, and suppose $s = d_x$. If we use the base loss \eqref{eq: base training loss} to train the student, the student can only see $\vbeta^{*T}\vx$, i.e. the action of every element in $\vbeta^* = (\beta^*_1, ..., \beta^*_4)$ on $\vx$ \textit{all at once}. On the other hand, as stated in the third bullet point of Theorem 3, for every $i\in\{1,...,s\}$, the $i^{\text{th}}$ hidden neuron $[\widetilde{\mW}_1]_{i,:}$ of $\widetilde{\mW}_1$ encodes exactly the $i^{\text{th}}$ entry in $\vbeta^*$, so the first group of the student neurons sees the action of $(\beta^*_1, \beta^*_2, 0, 0)$ on $\vx$, and the second group sees the action of $(0, 0, \beta^*_3, \beta^*_4)$ on $\vx$. In other words, the two groups of student neurons each observes response to the input $\vx$ created by a 2-sparse subset of $\vbeta^*$. Due to the lower sparsity in such responses, with the help of LASSO, the student neurons can learn more easily. 

On a more abstract level, the above theorem suggests an important angle of studying student-teacher learning: the ``simpler'' the hidden features of the teacher are, the more likely it is for the student to benefit from the teacher's features.

\subsection{Numerical evidence}
We verify our claims using a nonlinear network. 

\textbf{Network setting}. The networks are shallow and fully-connected, with $m=20,000$, and the activation function $\sigma(\cdot)$ is the ReLU function. We define $\ell(\cdot, \cdot)$ to be the square loss. All student networks are initialized with the Xavier Normal initialization, and optimized with SGD.

\textbf{Experiment setting}. The clean input signal $\vx\in\mathbb{R}^{500}$ has the distribution $\calN(\vzero, \mI)$, and the noise has distribution $\calN(\vzero, 0.09\mI)$. We assume that the ground truth network is identical to the teacher network. As a result, during testing, we simply compute $\E[(\widetilde{\mW}_2\mP\sigma(\mW_1(\vx+\vepsilon))-\widetilde{\mW}_2\mP\sigma(\widetilde{\mW}_1\vx))^2]$. To construct the teacher network, we set $\widetilde{\mW}_1$ with Xavier Normal initialization, and we set $[\widetilde{\mW}_2]_i = 1$ for all $i\in\{1, ..., m/2g\}$, and $[\widetilde{\mW}_2]_i = -1$ for all $i\in\{m/2g+1, ..., m/g\}$. Notice that, for any $g$ such that $m/g$ is divisible by $2$, the overall function $\widetilde{\mW}_2\mP\sigma(\widetilde{\mW}_1\cdot)$ remains the same, i.e. regardless of what $g$ is, a network trained with the base loss \eqref{eq: base training loss} remains the same. 

\begin{figure}[t!]
  \centering
  \begin{subfigure}
    \centering\includegraphics[width=1.0\linewidth]{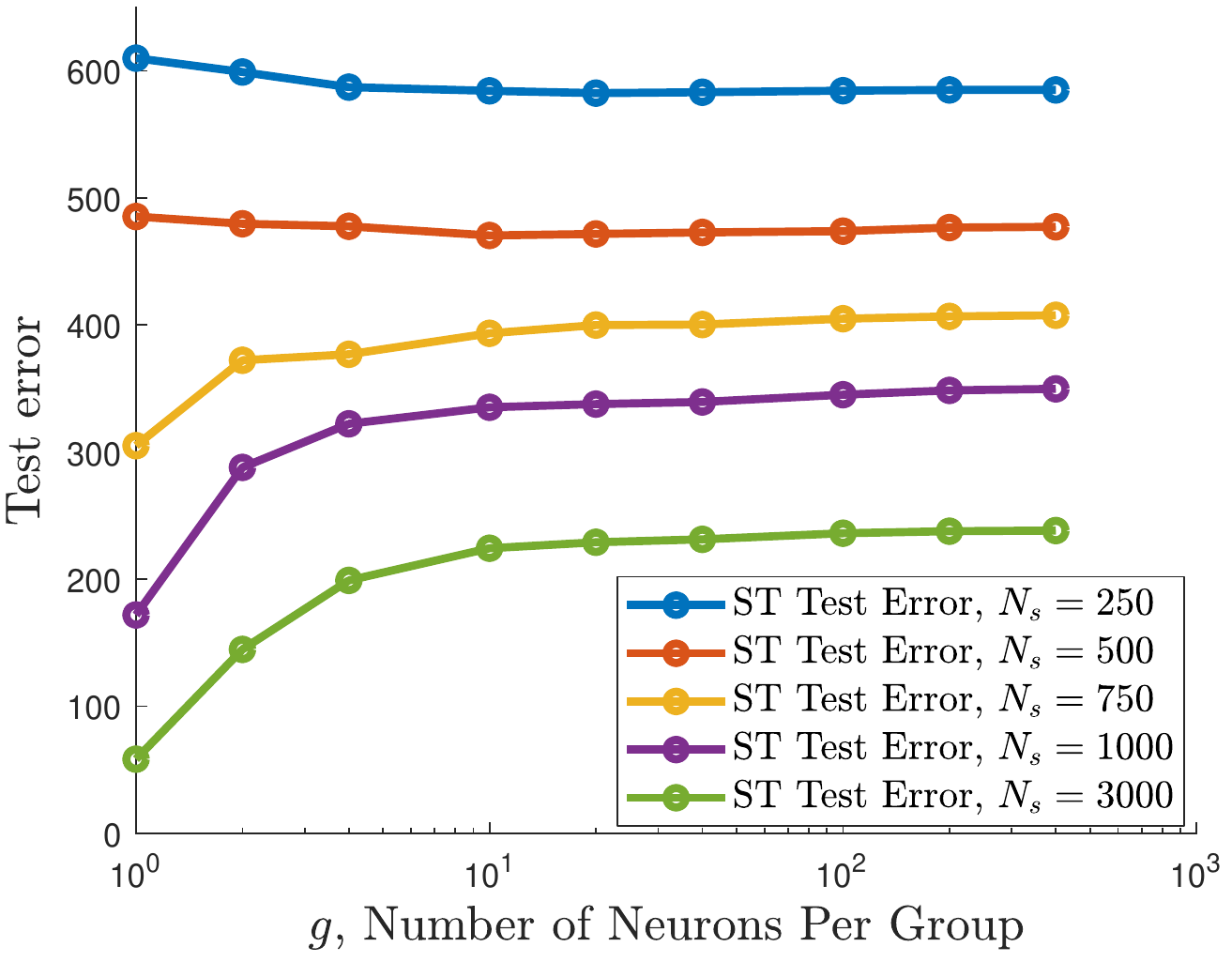}
  \end{subfigure}
  \caption{Here, $d_x = 500$ and $m=20,000$. Test error vs. $g$, the number of neurons per group. From the figure it is clear that, as long as $N_s$ is not too small, the \textit{fewer} neurons per group, the \textit{lower} the test error of the trained student network.}
  \label{fig: teacher decomp}
\end{figure}

\textbf{Interpreting the results}. As shown in Figure \ref{fig: teacher decomp}, as long as $N_s$ is not too small, the greater $g$ is, the higher the test error of the student trained with \eqref{eq: simplified ST loss}. Intuitively speaking, an increase in $g$ means that more teacher neurons are pooled in each of the $s/g$ groups, so the piecewise-linear function formed by each of these groups is more complex. Therefore, it becomes more difficult for the student's hidden neurons to learn with limited samples.

\section{Conclusion}
This paper offers a systematic analysis of the mechanism of feature-based student-teacher learning. Specifically, the ``when'' and ``why'' of the success of student-teacher learning in terms of generalization were studied. Through theoretical and numerical analysis, three conclusions were reached: use early stopping, use a knowledgeable teacher, and make sure that the teacher can decompose its hidden features well. It is our hope that the analytical and experimental results could help systematize the design principles of student-teacher learning, and potentially inspire new learning protocols that better utilize the hidden features of the teacher network, or construct networks that are better at ``teaching''.

\clearpage
\newpage
\nocite{buldygin_2000}
\nocite{eckart_1936}
\nocite{tropp_2006}
\nocite{cai_2011}
\nocite{wainwright_2009}
\nocite{vershynin_2018}
{\small
\bibliographystyle{ieee_fullname}
\bibliography{egbib}

\begin{thebibliography}{10}\itemsep=-1pt

\bibitem{Aguilar_2019}
Gustavo Aguilar, Yuan Ling, Yu Zhang, Benjamin Yao, Xing Fan, and Chenlei Guo.
\newblock Knowledge distillation from internal representations.
\newblock In {\em Proceedings of the AAAI Conference on Artificial
  Intelligence}, pages 34(05):7350--7357, 2020.

\bibitem{Arora_2018a}
Sanjeev Arora, Nadav Cohen, Noah Golowich, and Wei Hu.
\newblock A convergence analysis of gradient descent for deep linear neural
  networks.
\newblock In {\em arXiv preprint arXiv:1810.02281}, 2018.

\bibitem{Arora_2018b}
Sanjeev Arora, Nadav Cohen, and Elad Hazan.
\newblock On the optimization of deep networks: Implicit acceleration by
  overparameterization.
\newblock In {\em International Conference on Machine Learning (ICML)}, page
  244–253, 2018.

\bibitem{Arora_2019}
Sanjeev Arora, Nadav Cohen, Wei Hu, and Yuping Luo.
\newblock Implicit regularization in deep matrix factorization.
\newblock In {\em Advances in Neural Information Processing Systems (NIPS)},
  page 7413–7424, 2019.

\bibitem{Arora_2019b}
Sanjeev Arora, Simon~S. Du, Wei Hu, Zhiyuan Li, Ruslan Salakhutdinov, and
  Ruosong Wang.
\newblock On exact computation with an infinitely wide neural net.
\newblock In {\em Advances in Neural Information Processing Systems (NIPS)},
  volume~32, pages 8141--8150, 2019.

\bibitem{Buhlmann_2011}
Peter B\"uhlmann and Sara Van~De Geer.
\newblock {\em Statistics for high-dimensional data: methods, theory and
  applications}.
\newblock Springer, 2011.

\bibitem{buldygin_2000}
Valery~V. Buldygin and Kozachenko~Yu V.
\newblock {\em Metric Characterization of Random Variables and Random
  Processes}.
\newblock American Mathematical Society, 2000.

\bibitem{cai_2011}
T.~Tony Cai and Tiefeng Jiang.
\newblock Limiting laws of coherence of random matrices with applications to
  testing covariance structure and construction of compressed sensing matrices.
\newblock {\em Annals of Statistics}, 39(3):1496--1525, 06 2011.

\bibitem{Chi_2020}
Yiheng Chi, Abhiram Gnanasambandam, Vladlen Koltun, and Stanley~H. Chann.
\newblock Dynamic low-light imaging with quanta image sensors.
\newblock In {\em 16th European Conference on Computer Vision (ECCV)}, 2020.

\bibitem{eckart_1936}
Carl Eckart and Gale Young.
\newblock The approximation of one matrix by another of lower rank.
\newblock {\em Psychometrika}, 1:211--218, 1936.

\bibitem{Bengio_2010}
Xavier Glorot and Yoshua Bengio.
\newblock Understanding the difficulty of training deep feedforward neural
  networks.
\newblock In {\em Proceedings of AISTATS 2010}, volume~9, page 249–256, May
  2010.

\bibitem{Gnanasambandam_2020a}
Abhiram Gnanasambandam and Stanley~H. Chan.
\newblock Image classification in the dark using quanta image sensors.
\newblock In {\em 16th European Conference on Computer Vision (ECCV)}, 2020.

\bibitem{Gou_2020}
Jianping Gou, Baosheng Yu, Stephen~John Maybank, and Dacheng Tao.
\newblock Knowledge distillation: A survey.
\newblock {\em arXiv preprint arXiv:2006.05525}, 2020.

\bibitem{Hastie_2015}
Trevor Hastie, Robert Tibshirani, and Martin Wainwright.
\newblock {\em Statistical Learning with Sparsity: The Lasso and
  Generalizations}.
\newblock Chapman \& Hall/CRC, 2015.

\bibitem{He_2015}
Kaiming He, Xiangyu Zhang, Shaoqing Ren, and Jian Sun.
\newblock Delving deep into rectifiers: Surpassing human-level performance on
  imagenet classification.
\newblock In {\em Proceedings of the IEEE International Conference on Computer
  Vision (ICCV)}, pages 8141--8150, 2015.

\bibitem{Heo_2019}
Byeongho Heo, Jeesoo Kim, Sangdoo Yun, Hyojin Park, Nojun Kwak, and Jin~Young
  Choi.
\newblock A comprehensive overhaul of feature distillation.
\newblock In {\em Proceedings of the IEEE International Conference on Computer
  Vision (ICCV)}, page 1921–1930, 2019.

\bibitem{Hinton_knowledge_distill_2015}
Geoffrey Hinton, Oriol Vinyals, and Jeff Dean.
\newblock Distilling the knowledge in a neural network.
\newblock In {\em NIPS Deep Learning and Representation Learning Workshop},
  2015.

\bibitem{Hong_2020}
Ming Hong, Yuan Xie, Cuihua Li, and Yanyun Qu.
\newblock Distilling image dehazing with heterogeneous task imitation.
\newblock In {\em Proceedings of the IEEE Conference on Computer Vision and
  Pattern Recognition (CVPR)}, pages 3462--3471, 2020.

\bibitem{Jacot_2018}
Arthur Jacot, Franck Gabriel, and Clément Hongler.
\newblock Neural tangent kernel: convergence and generalization in neural
  networks.
\newblock In {\em Advances in Neural Information Processing Systems (NIPS)},
  volume~31, pages 8571--8580, 2018.

\bibitem{Jin_2019}
Xiao Jin, Baoyun Peng, Yichao Wu, Yu Liu, Jiaheng Liu, Ding Liang, Junjie Yan,
  and Xiaolin Hu.
\newblock Knowledge distillation via route constrained optimization.
\newblock In {\em Proceedings of the IEEE International Conference on Computer
  Vision (ICCV)}, pages 1345--1354, 2019.

\bibitem{Kawaguchi_2016}
Kenji Kawaguchi.
\newblock Deep learning without poor local minima.
\newblock In {\em Advances In Neural Information Processing Systems}, page
  586–594, 2016.

\bibitem{Kim_2018}
Jangho Kim, Seonguk Park, and Nojun Kwak.
\newblock Paraphrasing complex network: Network compression via factor
  transfer.
\newblock In {\em Advances in Neural Information Processing Systems (NIPS)},
  page 2760–2769, 2018.

\bibitem{liu_2020}
Pengpeng Liu, Irwin King, Michael~R. Lyu, and Jia Xu.
\newblock Flow2stereo: Effective self-supervised learning of optical flow and
  stereo matching.
\newblock In {\em Proceedings of the IEEE/CVF Conference on Computer Vision and
  Pattern Recognition (CVPR)}, June 2020.

\bibitem{Meinshausen_2006}
Nicolai Meinshausen and Bin Yu.
\newblock Lasso-type recovery of sparse representations for high-dimensional
  data.
\newblock Technical report, Departement of Statisics, UC Berkeley, 2006.

\bibitem{Phuong_2019}
Mary Phuong and Christoph Lampert.
\newblock Towards understanding knowledge distillation.
\newblock In {\em Proceedings of the 36th International Conference on Machine
  Learning, PMLR}, pages 97:5142--5151, 2019.

\bibitem{Rahbar_2020}
Arman Rahbar, Ashkan Panahi, Chiranjib Bhattacharyya, Devdatt Dubhashi, and
  Morteza~Haghir Chehreghani.
\newblock On the unreasonable effectiveness of knowledge distillation: Analysis
  in the kernel regime.
\newblock {\em arXiv preprint arXiv:2003.13438}, 2020.

\bibitem{Romero15-iclr}
Adriana Romero, Nicolas Ballas, Samira~Ebrahimi Kahou, Antoine Chassang, Carlo
  Gatta, and Yoshua Bengio.
\newblock Fitnets: Hints for thin deep nets.
\newblock In {\em Proceedings of the International Conference on Learning
  Representations (ICLR)}, 2015.

\bibitem{Saxe_2014}
Andrew~M. Saxe, James~L. McClelland, and Surya Ganguli.
\newblock Exact solutions to the nonlinear dynamics of learning in deep linear
  neural networks.
\newblock In {\em Proceedings of the International Conference on Learning
  Representations (ICLR)}, 2014.

\bibitem{schwartz2021isp}
Eli Schwartz, Alex Bronstein, and Raja Giryes.
\newblock Isp distillation, 2021.

\bibitem{Srinivas_2018}
Suraj Srinivas and Francois Fleuret.
\newblock Knowledge transfer with jacobian matching.
\newblock In {\em Proceedings of the 35th International Conference on Machine
  Learning, PMLR}, pages 80:4723--4731, 2018.

\bibitem{tropp_2006}
Joel~A. Tropp.
\newblock Just relax: convex programming methods for identifying sparse signals
  in noise.
\newblock {\em IEEE Transactions on Information Theory}, 52(3):1030--1051,
  2006.

\bibitem{Tung_2019}
Frederick Tung and Greg Mori.
\newblock Similarity-preserving knowledge distillation.
\newblock In {\em Proceedings of the IEEE International Conference on Computer
  Vision (ICCV)}, page 1365–1374, 2019.

\bibitem{Vapnik_2015}
Vladimir Vapnik and Rauf Izmailov.
\newblock Learning using privileged information: Similarity control and
  knowledge transfer.
\newblock In {\em Journal of Machine Learning Research}, page
  16(61):2023−2049, 2015.

\bibitem{vershynin_2018}
Roman Vershynin.
\newblock {\em High-Dimensional Probability: An Introduction with Applications
  in Data Science}.
\newblock Cambridge Series in Statistical and Probabilistic Mathematics.
  Cambridge University Press, 2018.

\bibitem{wainwright_2009}
M.~J. {Wainwright}.
\newblock Sharp thresholds for high-dimensional and noisy sparsity recovery
  using $\ell _{1}$ -constrained quadratic programming (lasso).
\newblock {\em IEEE Transactions on Information Theory}, 55(5):2183--2202,
  2009.

\bibitem{Wang_2020a}
Lin Wang and Kuk-Jin Yoon.
\newblock Knowledge distillation and student-teacher learning for visual
  intelligence: A review and new outlooks.
\newblock {\em arXiv preprint arXiv:2004.05937}, 2020.

\bibitem{Wang_2019}
Tao Wang, Li Yuan, Xiaopeng Zhang, and Jiashi Feng.
\newblock Distilling object detectors with fine-grained feature imitation.
\newblock In {\em Proceedings of the IEEE Conference on Computer Vision and
  Pattern Recognition (CVPR)}, page 4933–4942, 2019.

\bibitem{Wang_2017}
Wenhui Wang, Furu Wei, Li Dong, Hangbo Bao, Nan Yang, and Ming Zhou.
\newblock Minilm: Deep self-attention distillation for task-agnostic
  compression of pre-trained transformers.
\newblock In {\em arXiv preprint arXiv:2002.10957}, 2020.

\bibitem{Yim_2017}
Junho Yim, Donggyu Joo, Jihoon Bae, and Junmo Kim.
\newblock A gift from knowledge distillation: Fast optimization, network
  minimization and transfer learning.
\newblock In {\em Proceedings of the IEEE Conference on Computer Vision and
  Pattern Recognition (CVPR)}, page 4133–4141, 2017.

\end{thebibliography}
}

\onecolumn
\begin{center}
    \Huge Supplementary Materials
\end{center}
\tableofcontents

\addtocontents{toc}{\protect\setcounter{tocdepth}{2}}
\setcounter{section}{0}
\section{Introduction}
This document contains the supplementary materials to the paper ``Student-Teacher Learning from Clean Inputs to Noisy Inputs''. We shall provide the detailed versions of the theorems in the paper and their proofs. We will also provide some extra experimental results demonstrating the utility of student-teacher learning for the $\ell_1$-regularized linear networks, under the setting of section 5 of the paper.

\section{Proofs for Theorems in Section 3 of the Paper}
In this section, we shall present the proof for the Section 3 of the paper.

\subsection{Notations and Conventions}
Consider input-output training data pairs $\{(\vx_i, \vy_i)\}_{i = 1}^{N_s} \subset \mathbb{R}^{d_x}\times \mathbb{R}^{d_y}$, where $\vx_i$ is the $i$-th clean training sample, and $\vy_i$ is the $i$-th target. $\{\vepsilon_i\}_{i=1}^{N_s}\subset \mathbb{R}^{d_x}$ are the noise samples.

We write $\mX\in\mathbb{R}^{d_x \times N_s}$ as the clean input training data matrix, with its columns beings the $\vx_i$'s. Similarly, we construct the noisy input matrix $\mX_{\vepsilon}\in\mathbb{R}^{d_x \times N_s}$ and target matrix $\mY\in\mathbb{R}^{d_y \times N_s}$.

Given matrix $\mM$, we use $\text{row}(\mM)$ and $\text{col}(\mM)$ to denote the row and column spaces of matrix $\mM$. We use $\text{rank}(\mM)$ to denote the rank of the matrix. We use $\mP_{\mM}$ to denote the orthogonal projection matrix onto $\text{col}(\mM)$, and $\mP_{\mM}^{\perp}$ for projecting onto $\text{col}(\mM)^{\perp}$, the orthogonal complement of $\text{col}(\mM)$. We use $[\mM]_{i,j}$ to denote the $(i,j)$ entry in $\mM$. If $\mM\in\mathbb{R}^{n\times n}$ is symmetric, for its eigen-decomposition $\mM = \mU\mLambda\mU^T$, we assume that $[\mLambda]_{1,1}\ge [\mLambda]_{2,2}\ge ... \ge [\mLambda]_{n, n}$.

We consider a general deep linear network 
\begin{equation}
    \mW_{\textbf{L}} = \mW_L \mW_{L-1} ... \mW_1
\end{equation}
where $\mW_i \in \mathbb{R}^{d_i \times d_{i-1}}$. Set $d_0 \coloneqq d_x$ and $d_L \coloneqq d_y$. We restrict $L \ge 2$.

We denote $p \coloneqq \min_{i\in\{0, ..., L\}}d_i$. For any $(\mW_L, ..., \mW_1)$, clearly $\text{rank}(\mW_{\textbf{L}}) \le p$. We allow the networks to be wide, hence $\min_{i\in\{0, ..., L\}}d_i = \min(d_x, d_y)$ is possible.

For convenience, we will sometimes write $\mW_{i:j} = \mW_i\mW_{i-1}...\mW_j$. \textbf{Caution}: do not confuse this with the matrix notation $[\mW]_{i,j}$.

\subsection{Training Losses}
We consider two losses specialized to the deep linear networks.

The base loss (\textbf{we assume that it is the MSE loss in this whole section}):
\begin{equation}
\begin{aligned}
    (\mW_L^{\text{base}}, ..., \mW_1^{\text{base}})
    = & \argmin{\mW_L, ..., \mW_1} \widehat{\calL}_{\text{base}}(\mW_L, ..., \mW_1) \\
    = & \argmin{\mW_L, ..., \mW_1}\|\mW_{\textbf{L}}\mX_{\vepsilon} - \mY\|_F^2
    \label{eq: MSE loss, deep}
\end{aligned}
\end{equation}
To define the student-teacher loss (ST loss), first pick an $i^*\in\{1, ..., L\}$ (we exclude the trivial case $i^*=0$), and then define
\begin{equation}
\begin{aligned}
    (\mW_L^{\text{st}}, ..., \mW_1^{\text{st}})
    = & \argmin{\mW_L, ..., \mW_1}\widehat{\calL}_{\text{st}}(\mW_L, ..., \mW_1) \\
    = & \argmin{\mW_L, ..., \mW_1} \left(\|\mW_{\textbf{L}} \mX_{\vepsilon} - \mY \|_F^2 + \lambda \|\mW_{i^*:1}\mX_{\vepsilon} -\widetilde{\mW}_{i^*:1}\mX\|_F^2\right)
    \label{eq: SnT loss, deep}
\end{aligned}
\end{equation}
where we use the tuple $(\widetilde{\mW}_L, ..., \widetilde{\mW}_1)$ to denote the teacher's weight. Recall that \textit{the student and teacher share the same architecture}.

\subsection{Proof of Theorem 1 from Paper}
In this subsection, we prove theorem 1 from the paper, i.e. we focus on the undersampling regime $N_s < d_x$. Moreover, we assume that $L=2$ for the teacher and student. In this case, the hidden dimension of the networks is just $d_1$.

Recall from the paper that we assume the base loss is MSE.

We first restate the theorem from the paper, with all assumptions precisely described.

\begin{theorem}[Theorem 1 from paper, detailed version]
Denote $\mW^{\text{base}}_i(t)$ and $\mW^{\text{st}}_{i}(t)$ as the weights for the student network during training with the the base loss \eqref{eq: MSE loss, deep} and the student-teacher loss \eqref{eq: SnT loss, deep}, respectively.

Let the following assumptions hold:
\begin{enumerate}
    \item The optimizer is gradient flow;
    \item $N_s < d_x$;
    \item $L=2$;
    \item $\{(\vx_i,\vy_i)\}_{i=1}^{N_s}$ and $\{\vepsilon_i\}_{i=1}^{N_s}$ are all sampled independently, and $\vx$ and $\vepsilon$ are continuous random vectors;
    \item There exists some $\delta > 0$ such that $\|\mW_i^{\text{base}}(0)\|_F \le \delta$ and $\|\mW_i^{\text{st}}(0)\|_F \le \delta$ for all $i$;
    \item The teacher network $(\widetilde{\mW}_2, \widetilde{\mW}_1)$ minimizes the training loss for clean data: 
    \begin{equation}
        (\widetilde{\mW}_2, \widetilde{\mW}_1) = \argmin{\mW_2, \mW_1} \widehat{\calL}_{teacher}(\mW_2, \mW_1)= \argmin{\mW_2, \mW_1}\|\mW_{\textbf{L}}\mX - \mY\|_F^2
        \label{eq: teacher loss, deep}
    \end{equation}
    \item The $\mW^{\text{base}}_i(0)$'s are initialized with the balanced initialization \cite{Arora_2018b}, i.e.
    \begin{equation}
        \mW^{\text{base}}_2(0)^T\mW^{\text{base}}_2(0) = \mW^{\text{base}}_1(0)\mW^{\text{base}}_1(0)^T
    \end{equation}
    \item The gradient flow successfully converges to a global minimizer for both the MSE- and student-teacher-trained networks;
    \item The weights $\mW_i^{\text{st}}(t)$ remain in a compact set for $t\in[0,\infty)$. In particular, denote $\|\mW_i^{\text{st}}(t)\|_F \le M, t\in[0,\infty)$.
\end{enumerate}

When $\delta$ is sufficiently small, the following is true almost surely:
\begin{equation}
    \lim_{t\to\infty} \|\mW_{\textbf{L}}^{\text{base}}(t) - \mW_{\textbf{L}}^{\text{st}}(t)\|_F \le C\delta
\end{equation}
where $C$ is a constant independent of $\delta$.
\end{theorem}
\begin{proof}
By lemma \ref{lemma: theorem 1, MSE grad flow} and lemma \ref{lemma: theorem 1, ST grad flow} below, and applying the triangle inequality, we obtain
\begin{equation}
    \lim_{t\to\infty}\|\mW_{\textbf{L}}^{\text{base}}(t) - \mW_{\textbf{L}}^{\text{st}}(t)\|_F \le C \delta
\end{equation}
where $C\in\calO(M+p^{1/4}\|\mU_{p}\mU_{p}^T\mY(\mX_{\vepsilon}^T\mX_{\vepsilon})^{-1}\mX_{\vepsilon}^T\|_F^{1/2})$ when $\delta$ is sufficiently small ($\mU_p$ shall be defined below).
\end{proof}

\subsubsection{Main Lemmas}
We define and elaborate on some terms that will be used frequently throughout this subsection.

First recall that $p\coloneqq \min(d_x, d_1, d_y)$. We define the matrix $\mU_p\in\mathbb{R}^{d_y\times p}$ as follows. The columns of $\mU_{p}$ are the dominant $p$ eigenvectors of the matrix $\mY\mY^T = \mU\mLambda\mU^T$ (assuming that the eigenvalues in all the eigen-decompositions are sorted from largest to smallest). Note that if $\text{rank}(\mY) < p$, then one can choose arbitrary unit vectors orthogonal to the dominant $\text{rank}(\mY)$ eigenvectors of $\mY\mY^T$ as the last $p - \text{rank}(\mY)$ columns in $\mU_{p}$.

\begin{lemma}[Bias of MSE-induced Gradient Flow]
\label{lemma: theorem 1, MSE grad flow}
With the assumptions in the main theorem, the following holds almost surely:
\begin{equation}
    \lim_{t\to\infty} \mW_{\textbf{L}}^{\text{base}}(t) = \mU_p\mU_p^T\mY(\mX_{\vepsilon}^T\mX_{\vepsilon})^{-1}\mX_{\vepsilon}^T + \mW(\delta)
\end{equation}
where $\|\mW(\delta)\|_F \le C\delta$, for some $C \in\mathcal{O}(p^{1/4}\gamma^{1/2})$, when $\delta$ is sufficiently small, and $\gamma \coloneqq \|\mU_{p}\mU_{p}^T\mY(\mX_{\vepsilon}^T\mX_{\vepsilon})^{-1}\mX_{\vepsilon}^T\|_F$. 
\end{lemma}
\begin{proof}
In this proof, for the sake of readability, we will write $\mW_i(t) = \mW^{\text{base}}_i(t)$. We also abuse notation a bit by writing $\mW_i(\infty)$, with the understanding that they mean $\lim_{t\to\infty}\mW_i(t)$. These limits do exist, due to our assumption that gradient flow converges to a global minimizer.

The proof has three steps:
\begin{enumerate}
    \item \textbf{Structure of the Solution}.
    
    We prove that 
    \begin{equation}
        \mW_2(\infty)\mW_1(\infty) = \mU_{p}\mU_{p}^T\mY(\mX_{\vepsilon}^T\mX_{\vepsilon})^{-1}\mX_{\vepsilon}^T + \mW_2(\infty)\mW_1(0)_{\perp}
    \end{equation}
    where we orthogonally decomposed the row space $\mW_1(0) = \mW_1(0)_{\parallel} + \mW_1(0)_{\perp}$, where $\text{row}(\mW_1(0)_{\parallel})\subseteq\text{col}(\mX_{\vepsilon})$ and $\text{row}(\mW_1(0)_{\perp})\subseteq\text{col}(\mX_{\vepsilon})^{\perp}$.
    
    We begin by observing the updates made by gradient flow to $\mW_1$:
    
    \begin{equation}
    \begin{aligned}
        \frac{\partial \mW_1}{\partial t} = & \eta(\mW_2(t)^T\mY\mX_{\vepsilon}^T - \mW_2(t)^T\mW_2(t)\mW_1(t)\mX_{\vepsilon}\mX_{\vepsilon}^T).
    \end{aligned}
    \end{equation}
    Here, $\eta$ is the update step size, and assumed to be close to $0$. As explained in \cite{Arora_2018b} 
    section 5, when $\eta^2\approx 0$, the discrete gradient descent steps translate into the gradient flow differential equation. The right-hand side of this differential equation is simply the derivative of the base MSE loss \eqref{eq: MSE loss, deep} with respect to $\mW_1$.
    
    Notice that $\text{row}\left(\frac{\partial \mW_1}{\partial t}\right)\subseteq \text{col}(\mX_{\vepsilon})$ at all time. We have the following:
    \begin{equation}
        \text{row}(\mW_1(\infty) - \mW_1(0)) = \text{row}\left(\int_{t=0}^{\infty}\frac{\partial \mW_1}{\partial t}dt\right) \subseteq \text{col}(\mX_{\vepsilon}).
    \end{equation}
    The infinite integral is well-defined since we assumed the convergence of gradient flow. The above observation, combined with our definition of $\mW_1(0)_{\parallel}$ and $\mW_1(0)_{\perp}$ from before, imply that gradient flow only modifies $\mW_1(0)_{\parallel}$, and leaves the $\mW_1(0)_{\perp}$ untouched. In other words, decomposing the row vectors of $\mW_1(\infty)$ orthogonally w.r.t $\text{col}(\mX_{\vepsilon})$ (identical to what we did with $\mW_1(0)$), we can write
    \begin{equation}
        \mW_1(\infty) = \mW_1(\infty)_{\parallel} + \mW_1(\infty)_{\perp} = \mW_1(\infty)_{\parallel} + \mW_1(0)_{\perp}.
    \end{equation}
    
    The important point to notice is that, 
    \begin{equation}
        \mW_2(\infty)\mW_1(\infty) = \mW_2(\infty)\mW_1(\infty)_{\parallel} + \mW_1(\infty)\mW_1(0)_{\perp}.
    \end{equation}
    
    Recalling the expression of global minimizers stated in Lemma \ref{lemma: glob min, MSE, N < d_x}, with probability $1$ (over the randomness in the training sample matrix $\mX_{\epsilon}$), all the global minimizers share exactly the same structure as we have for $\mW_2(\infty)\mW_1(\infty)$, i.e. these minimizers consist of two terms, first, the minimum-Frobenius-norm solution $\mU_{p}\mU_{p}^T\mY(\mX_{\vepsilon}^T\mX_{\vepsilon})^{-1}\mX_{\vepsilon}^T$ whose row space lies in $\text{col}(\mX_{\vepsilon})$, and second, the ``residue matrix'' $\mR$ whose row space lies in $\text{col}(\mX_{\vepsilon})^{\perp}$. It follows that $\mW_2(\infty)\mW_1(\infty)_{\parallel} = \mU_{p}\mU_{p}^T\mY(\mX_{\vepsilon}^T\mX_{\vepsilon})^{-1}\mX_{\vepsilon}^T$, which finishes the first step of the overall proof.
    
    \item \textbf{Uniform Upper Bound on $\|\mW_2(\infty)\|_F$ for Small and Balanced Initialization}.
    
    We relate $\|\mW_2(\infty)\|_F$ to $\|\mW_2(\infty)\mW_1(\infty)\|_F$.
    
    Let's denote the SVDs of $\mW_2(\infty) = \mU^{(2)}\mLambda^{(2)}\mV^{(2)T}$, and $\mW_1(\infty) = \mU^{(1)}\mLambda^{(1)}\mV^{(1)T}$.
    
    A deep linear network that is initialized in the balanced fashion remains balanced throughout training \cite{Arora_2018b} (theorem 1), therefore, we have that $\mW_2^T(\infty)\mW_2(\infty) = \mW_1(\infty)\mW_1^T(\infty)$, which means that
    \begin{equation}
        \mV^{(2)}\mLambda^{(2)T}\mLambda^{(2)}\mV^{(2)T} = \mU^{(1)}\mLambda^{(1)}\mLambda^{(1)T}\mU^{(1)T}.
    \end{equation}
    In other words, $\mLambda^{(2)T}\mLambda^{(2)} = \mLambda^{(1)}\mLambda^{(1)T}$, i.e. $[\mLambda^{(2)}]_{i,i} = [\mLambda^{(1)}]_{i,i}$ for $i \in \{1, ..., d_1\}$, and the orthogonal matrices $\mV^{(2)}$ and $\mU^{(1)}$ are equal up to some rotation in the eigenspaces corresponding to each eigenvalue in $\mLambda^{(2)T}\mLambda^{(2)}$ (see the details in \cite{Arora_2018b} Appendix A.1). It also follows that, $\text{rank}(\mLambda^{(1)}) = \text{rank}(\mLambda^{(2)}) \le p = \min(d_x, d_1, d_y)$. Using equations (23) and (24) (and the equations before these two) from \cite{Arora_2018b}, it follows that 
    \begin{equation}
    \begin{aligned}
        \|\mW_2(\infty)\mW_1(\infty)\|_F
        = & \|\mLambda^{(2)}\mLambda^{(2)T}\|_F \\
        = & \sqrt{\sum_{i=1}^{p} [\mLambda^{(2)}]_{i,i}^4}.
    \end{aligned}
    \end{equation}
    Recall that, by H\"older's inequality, $\|\vx\|_1 \le \sqrt{p}\|\vx\|_2$  for any $\vx\in\mathbb{R}^{p}$. Therefore
    \begin{equation}
        \|\mW_2(\infty)\|_F^2 = \sum_{i=1}^{p} [\mLambda^{(2)}]_{i,i}^2 \le \sqrt{p}\sqrt{\sum_{i=1}^{p} [\mLambda^{(2)}]_{i,i}^4} = \sqrt{p}\|\mW_2(\infty)\mW_1(\infty)\|_F.
    \end{equation}
    
    Let's study the term $\|\mW_2(\infty)\mW_1(\infty)\|_F$. By the Pythagorean theorem, $\|\mW_2(\infty)\mW_1(\infty)_{\parallel}\|_F^2 +  \|\mW_2(\infty)\mW_1(0)_{\perp}\|_F^2 = \|\mW_2(\infty)\mW_1(\infty)\|_F^2$. Since $\|\mW_1(0)\|_F = \delta$, and recalling the definition $\gamma \coloneqq \|\mU_{p}\mU_{p}^T\mY(\mX_{\vepsilon}^T\mX_{\vepsilon})^{-1}\mX_{\vepsilon}^T\|_F$, we have
    \begin{equation}
    \begin{aligned}
        & \|\mW_2(\infty)\mW_1(\infty)\|_F^2 \le \gamma^2 + \delta^2\|\mW_2(\infty)\|_F^2 \\
        \implies & \|\mW_2(\infty)\mW_1(\infty)\|_F \le \sqrt{\gamma^2 + \delta^2\|\mW_2(\infty)\|_F^2} < \gamma + \delta\|\mW_2(\infty)\|_F.
    \end{aligned}
    \end{equation}
    
    Therefore,
    \begin{equation}
    \begin{aligned}
        & \|\mW_2(\infty)\|_F^2 < \sqrt{p}\gamma + \sqrt{p}\delta\|\mW_2(\infty)\|_F \\
        \iff & \|\mW_2(\infty)\|_F^2 - \sqrt{p}\delta\|\mW_2(\infty)\|_F - \sqrt{p}\gamma < 0 \\
        \iff & \frac{\sqrt{p}\delta - \sqrt{p\delta^2 + 4\sqrt{p}\gamma}}{2} < \|\mW_2(\infty)\|_F < \frac{\sqrt{p}\delta + \sqrt{p\delta^2 + 4\sqrt{p}\gamma}}{2} \\
        \implies & \|\mW_2(\infty)\|_F < \frac{\sqrt{p}\delta + \sqrt{p\delta^2 + 4\sqrt{p}\gamma}}{2} < \sqrt{p}\delta + p^{1/4}\gamma^{1/2}.
    \end{aligned}
    \end{equation}
    The upper bound is clearly $\mathcal{O}(p^{1/4}\gamma^{1/2})$ for $\delta$ sufficiently small. 
    
    \item \textbf{Conclusion}.
    
    The desired result now follows by combining 1. and 2., and by applying Cauchy-Schwartz to $\|\mW_2(\infty)\mW_1(0)_{\perp}\|_F$, with $C = \sqrt{p}\delta + p^{1/4}\gamma^{1/2}$.
\end{enumerate}
\end{proof}

\begin{lemma}[Bias of Student-teacher-induced Gradient Flow]
\label{lemma: theorem 1, ST grad flow}
With the assumptions in the main theorem, the following holds:
\begin{equation}
    \lim_{t\to\infty}\mW_{\textbf{L}}^{\text{st}}(t) = \mU_{p}\mU_{p}^T\mY(\mX_{\vepsilon}^T\mX_{\vepsilon})^{-1}\mX_{\vepsilon}^T + \mW(\delta)
\end{equation}
where $\|\mW(\delta)\|_F \le M\delta$; recall that we assumed $\|\mW_i^{\text{st}}(t)\|\le M$ for all $t\in[0,\infty)$. In other words, small initialization leads to $\lim_{t\to\infty}\mW_{\textbf{L}}(t) \approx \mU_{p}\mU_{p}^T\mY(\mX_{\vepsilon}^T\mX_{\vepsilon})^{-1}\mX_{\vepsilon}^T$.

\end{lemma}
\begin{proof}
We write $\mW_i(t) = \mW^{\text{st}}_i(t)$ for notational simplicity.

First observe that
\begin{equation}
\begin{aligned}
    \frac{\partial \mW_1}{\partial t} = & \eta\mW_2(t)^T(\mY - \mW_2(t)\mW_1(t)\mX_{\vepsilon})\mX_{\vepsilon}^T + \eta\lambda(\widetilde{\mW}_1\mX - \mW_1(t)\mX_{\vepsilon})\mX_{\vepsilon}^T.
\end{aligned}
\end{equation}
It follows that $\text{row}\left(\frac{\partial \mW_1}{\partial t}\right) \subseteq \text{col}(\mX_{\vepsilon})$. Therefore, arguing similarly to step 1 of the proof of lemma \ref{lemma: theorem 1, MSE grad flow}, we may write $\mW_1(t) = \mW_1(t)_{\parallel} + \mW_1(t)_{\perp}= \mW_1(t)_{\parallel} + \mW_1(0)_{\perp}$, where $\text{row}(\mW_1(t)_{\parallel})\subseteq\text{col}(\mX_{\vepsilon})$, and $\text{row}(\mW_1(0)_{\perp})\subseteq\text{col}(\mX_{\vepsilon})^{\perp}$.

Knowing the form of global minimizers of the ST loss from Lemma \ref{lemma: glob min, ST, N < d_x}, we know 
\begin{equation}
    \mW_{\textbf{L}}(\infty) = \mU_{p}\mU_{p}^T\mY(\mX_{\vepsilon}^T\mX_{\vepsilon})^{-1}\mX_{\vepsilon}^T + \mW_2(\infty)\mW_1(0)_{\perp}
\end{equation}

Therefore $\|\mW_{\textbf{L}}(\infty) - \mU_{p}\mU_{p}^T\mY(\mX_{\vepsilon}^T\mX_{\vepsilon})^{-1}\mX_{\vepsilon}^T\|_F \le M\delta$.
\end{proof}

\subsubsection{Auxilliary Lemmas}
\begin{lemma}[Global minimizers of MSE loss \eqref{eq: MSE loss, deep}, $N < d_x$]
\label{lemma: glob min, MSE, N < d_x}
The set of global minimizers to the MSE loss \eqref{eq: MSE loss, deep} is the following almost surely (over the randomness of the training samples):
\begin{equation}
    \{\mW_2,\mW_1| \; \mW_2\mW_1 = \mU_{p}\mU_{p}^T\mY(\mX_{\vepsilon}^T\mX_{\vepsilon})^{-1}\mX_{\vepsilon}^T + \mR, \; \text{row}(\mR) \subseteq \text{col}(\mX_{\vepsilon})^{\perp}\}
        \label{eq: MSE loss solution}
\end{equation}
where the columns of $\mU_{p}$ are the dominant $p$ eigenvectors of the matrix $\mY\mY^T = \mU\mLambda\mU^T$. Note that if $\text{rank}(\mY) < p$, then one can choose arbitrary unit vectors orthogonal to the dominant $\text{rank}(\mY)$ eigenvectors of $\mY\mY^T$ as the last $p - \text{rank}(\mY)$ columns in $\mU_{p}$.
\end{lemma}
\begin{proof}
First of all, note that since $\vx$ and $\vepsilon$ are continuous random vectors, $\mX_{\epsilon}$ must be full rank almost surely, so $(\mX_{\vepsilon}^T\mX_{\vepsilon})^{-1}$ exists.

Now, we note that
\begin{equation}
\begin{aligned}
    & \{\mW\in\mathbb{R}^{d_y\times d_x} | (\text{rank}(\mW)\le p) \;\wedge\; (\mW \text{ minimizes } \|\mW'\mX_{\epsilon}-\mY\|_F^2\} \\
    = & \{\mW_2\mW_1 | (\mW_2 \in \mathbb{R}^{d_y\times d_1}) \wedge (\mW_1\in\mathbb{R}^{d_1\times d_x}) \wedge ((\mW_2,\mW_1) \text{ minimizes \eqref{eq: MSE loss, deep}})\}.
\end{aligned}    
\end{equation}
To see ``$\subseteq$'' direction, take any $\mW$ in the first set, we can decompose it as $\mW = \mA_W\mB_W$ where $\mA_W\in\mathbb{R}^{d_y\times d_1}$ and $\mB_W\in\mathbb{R}^{d_1\times d_x}$, and $(\mA_W\mB_W)$ clearly minimizes \eqref{eq: MSE loss, deep}. The other direction can also be easily shown.

It follows that, the set of solutions $\mW_{\textbf{L}} = \mW_2\mW_1$ that we need to solve for is the same as the set
\begin{equation}
    \argmin{\text{rank}(\mW) \le d_1} \|\mY - \mW\mX_{\vepsilon}\|^2_F.
\end{equation}
This is basically a low-rank approximation problem. By the Eckart-Young-Mirsky theorem \cite{eckart_1936} [10], the matrix $\mU_{p}\mU_{p}^T\mY$ is the best approximation to the matrix $\mY$ under the Frobenius norm with rank no greater than $p$.

To achieve this solution, we need
\begin{equation}
\begin{aligned}
    & \mW\mX_{\vepsilon} = \mU_{p}\mU_{p}^T\mY \\
    \iff & \mW\mX_{\vepsilon} = \mU_{p}\mU_{p}^T\mY(\mX_{\vepsilon}^T\mX_{\vepsilon})^{-1}\mX_{\vepsilon}^T\mX_{\vepsilon} \\
    \iff & (\mW - \mU_{p}\mU_{p}^T\mY(\mX_{\vepsilon}^T\mX_{\vepsilon})^{-1}\mX_{\vepsilon}^T)\mX_{\vepsilon} = \vzero \\
    \iff & \mW - \mU_{p}\mU_{p}^T\mY(\mX_{\vepsilon}^T\mX_{\vepsilon})^{-1}\mX_{\vepsilon}^T = \mR, \; \text{ s.t. } \text{row}(\mR) \subseteq \text{col}(\mX_{\vepsilon})^{\perp}.
    \label{derivation: MSE solution}
\end{aligned}
\end{equation}
which concludes the proof.
\end{proof}

\begin{lemma}[Global Minimizers of ST loss \eqref{eq: SnT loss, deep}, $N < d_x$]
\label{lemma: glob min, ST, N < d_x}
Choose a global minimizer $(\widetilde{\mW_2},\widetilde{\mW_1})$ of \eqref{eq: teacher loss, deep}. Then almost surely, the set of global minimizers to \eqref{eq: SnT loss, deep} is the following:
\begin{equation}
\begin{aligned}
    \{\mW_2\in\mathbb{R}^{d_y\times d_1}, \mW_1\in\mathbb{R}^{d_1\times d_x}| & \; \mW_2\mW_1 = \mU_{p}\mU_{p}^T\mY(\mX_{\vepsilon}^T\mX_{\vepsilon})^{-1}\mX_{\vepsilon}^T + (\widetilde{\mW}_2 + \mR_2)\mR_1, \\
    & \; \mR_1\in\mathbb{R}^{d_1\times d_x} \land \, \text{row}(\mR_1) \subseteq \text{col}(\mX_{\vepsilon})^{\perp} \land \, \mR_2\in\mathbb{R}^{d_x \times d_1} \land \, \text{row}(\mR_2) \subseteq \text{col}(\widetilde{\mW}_1\mX)^{\perp} \}.
    \label{eq: SnT solution}
\end{aligned}
\end{equation}
\end{lemma}
\begin{remark}
Notice that this solution set is just a subset of the MSE solution set from lemma \ref{lemma: glob min, MSE, N < d_x}. In the MSE solution set, the ``residue matrix'' $\mR$ just satisfies $\text{row}(\mR)\subseteq\text{col}(\mX_{\epsilon})^{\perp}$. For this ST loss solution set, the ``residue matrix'' also satisfies $\text{row}((\widetilde{\mW}_2 + \mR_2)\mR_1)\subseteq\text{col}(\mX_{\epsilon})^{\perp}$, although it does have more structure.
\end{remark}

\begin{proof}
Like in the last lemma, note that since $\vx$ and $\vepsilon$ are continuous random vectors, $\mX_{\epsilon}$ must be full rank almost surely, so $(\mX_{\vepsilon}^T\mX_{\vepsilon})^{-1}$ exists.

The proof relies on the fact that,
\begin{equation}
\begin{aligned}
    &\min_{\mW_2, \mW_1}\left\{ \|\mY - \mW_2\mW_1\mX_{\vepsilon}\|^2_F + \lambda \|\widetilde{\mW}_1\mX - \mW_1\mX_{\vepsilon}\|_F^2\right\} \\
    \ge & \min_{\mW_2,\mW_1} \left\{\|\mY - \mW_2\mW_1\mX_{\vepsilon}\|^2_F \right\} + \min_{\mW_1}\left\{\lambda \|\widetilde{\mW}_1\mX - \mW_1\mX_{\vepsilon}\|_F^2\right\}
\end{aligned}
\end{equation}
We can see that the lower bound is achievable only when we individually minimize the two loss terms in the lower bound, in other words, denoting $l_1(\mW_2, \mW_1) = \|\mY - \mW_2\mW_1\mX_{\vepsilon}\|^2_F$, and $l_2(\mW_1) = \|\widetilde{\mW}_1\mX - \mW_1\mX_{\vepsilon}\|_F^2$, equality is true only for $(\mW_2, \mW_1)$ that lies in the following intersection
\begin{equation}
    \{\mW_2,\mW_1 | \mW_2, \mW_1 \text{ minimizes } l_1 \} \cap \{\mW_1 | \mW_1 \text{ minimizes } l_2 \}
\end{equation}

We proceed to minimize the two terms individually. 

First notice that the regularizer can be made $0$ with the following set of expressions for $\mW_1$
\begin{equation}
    \{\mW_1 | \; \mW_1 = \widetilde{\mW}_1\mX(\mX_{\vepsilon}^T\mX_{\vepsilon})^{-1}\mX_{\vepsilon}^T + \mR_1, \; \mR_1\in\mathbb{R}^{d_1\times d_x} \land \, \text{row}(\mR_1) \subseteq \text{col}(\mX_{\vepsilon})^{\perp}\}
\end{equation}
Its proof is very similar to \eqref{derivation: MSE solution}.

We already know the set of minimizers of the MSE loss from the previous section. We take the intersection of the two sets. First note that $\mW_1\mX_{\vepsilon} = \widetilde{\mW}_1\mX$, therefore, for $\mW_2, \mW_1$ to minimize the MSE loss $l_1$, we need $\mW_2\widetilde{\mW}_1\mX = \mU_{p}\mU_{p}^T\mY$. But since $(\widetilde{\mW}_2, \widetilde{\mW}_1)$ minimizes \eqref{eq: teacher loss, deep}, $\widetilde{\mW}_2\widetilde{\mW}_1\mX = \mU_{p}\mU_{p}^T\mY$ has to be true (can be proven using essentially the same argument as in the previous lemma). In other words, 
\begin{equation}
    \mW_2\widetilde{\mW}_1\mX = \widetilde{\mW}_2\widetilde{\mW}_1\mX \iff (\mW_2 - \widetilde{\mW}_2)\widetilde{\mW}_1\mX = \vzero
\end{equation}
The rest of the proof follows directly from here.
\end{proof}

\subsection{Proof of Theorem 2 from Paper}

Similar to the proof for theorem 1 of the paper, we restate the theorem itself precisely first, and then present its proof and the relevant lemmas.

Again, recall that the base loss is MSE.

\begin{theorem} [Theorem 2 from paper, detailed version]
Assume the following:
\begin{enumerate}
    \item $N \ge d_x$, and $\mX_{\epsilon}$ is full rank;
    \item $L \ge 2$ (a general deep linear network);
    \item $p \coloneqq \min_{i\in\{0, ..., L\}}d_i \ge  \text{rank}(\mY\mX_{\vepsilon}^T(\mX_{\vepsilon}\mX_{\vepsilon}^T)^{-1})$
    \item $\widetilde{\mW}_{\textbf{L}}\mX = \mY$

\end{enumerate}

Then the global minimizers
\begin{equation}
    \mW_{\textbf{L}}^{\text{base}} = \mW_{\textbf{L}}^{\text{st}} = \mY\mX_{\vepsilon}^T(\mX_{\vepsilon}\mX_{\vepsilon}^T)^{-1}
\end{equation}
\end{theorem}
\begin{remark}
Let us interpret the assumptions.
\begin{itemize}
    \item Assumption 3. ensures that $\mW_{\textbf{L}}^{\text{base}} = \mY\mX_{\vepsilon}^T(\mX_{\vepsilon}\mX_{\vepsilon}^T)^{-1}$ can be true.
    
    Intuitively speaking, this assumption is requiring the student to be sufficiently ``complex'' for the task that it needs to solve.
    
    \item Assumption 4. enforces that the teacher perfectly interpolates the clean input-output training pairs. 
    
    This assumption can be satisfied by enforcing, for instance, that $\text{rank}(\mX) \le d_x$, and the maximum possible rank of $\widetilde{\mW}_{\textbf{L}}$ is no less than $\text{rank}\left(\mY\left(\overline{\mX}^T\overline{\mX}\right)^{-1}\overline{\mX}^T\right)$, where $\overline{\mX}$ is constructed by removing every linearly dependent column of $\mX$. 
    
    Intuitively speaking, we are requiring that the task which the teacher needs to solve is sufficiently ``simple''.
    
    \item If a slightly stronger condition was added, the argument in our proof can in fact handle the situation that the teacher and the student networks have different architectures, and the only requirements on the teacher are that, its hidden feature's dimension matches $d_{i^*}$, and the teacher can perfectly interpolate the clean training samples.
    
\end{itemize}

\end{remark}
\begin{proof}
This proof is divided into two parts. We study the MSE and student-teacher solutions respectively.

\begin{enumerate}
    \item We first study the global minimizers of the MSE loss. First of all, the following is true:
    \begin{equation}
    \begin{aligned}
        & \{\mW_{\textbf{L}}| (\mW_L, ..., \mW_1) \text{ minimizes } \eqref{eq: MSE loss, deep})\} \\
        = &\left\{\mW \Big|\mW = \argmin{\text{rank}(\mW) \le p} \|\mW\mX_{\vepsilon} - \mY\|_F^2\right\}
    \end{aligned}
    \end{equation}
    To see the ``$\subseteq$'' direction, take any $(\mW_L, ..., \mW_1)$ that minimizes the MSE loss \eqref{eq: MSE loss, deep}, clearly $\text{rank}(\mW_{\textbf{L}}) \le p$. Furthermore, $\mW_{\textbf{L}}$ must minimize the single-layer-network rank-restricted MSE loss, since if it was not true, then there exists some $\mW^*$ with $\text{rank}(\mW^*)\le p$ such that
    \begin{equation}
        \|\mW^*\mX_{\vepsilon} - \mY\|_F^2 < \|\mW_{\textbf{L}}\mX_{\vepsilon} - \mY\|_F^2
    \end{equation}
    But clearly one can find a tuple $(\mW^*_L, ..., \mW^*_1)$ that decomposes $\mW^*$, which contradicts the minimality of $(\mW_L, ..., \mW_1)$. The ``$\supseteq$'' direction can be proven in a similar way.
    
    Therefore, it suffices to study the set of global minimizers of the rank-restricted MSE problem 
    \begin{equation}
        \overline{\mW} = \argmin{\text{rank}(\mW) \le p}\|\mW\mX_{\vepsilon} - \mY\|_F^2
    \end{equation}
    
    With our assumption that $p \ge \text{rank}(\mY\mX_{\vepsilon}^T(\mX_{\vepsilon}\mX_{\vepsilon}^T)^{-1})$, clearly $\overline{\mW}$ is unique and $\overline{\mW} = \mY\mX_{\vepsilon}^T(\mX_{\vepsilon}\mX_{\vepsilon}^T)^{-1}$. It follows that 
    \begin{equation}
        \mW^{\text{base}}_{\textbf{L}} = \mY\mX_{\vepsilon}^T(\mX_{\vepsilon}\mX_{\vepsilon}^T)^{-1}
    \end{equation}

    \item We now study the student-teacher loss \eqref{eq: SnT loss, deep}. 
    
    Note the inequality
    \begin{equation}
        \min_{(\mW_L, ..., \mW_1)} \widehat{\calL}_{\text{st}}(\mW_L, ..., \mW_1) \ge \min_{(\mW_L, ..., \mW_1)} \widehat{\calL}_{\text{base}}(\mW_L, ..., \mW_1) + \lambda \min_{(\mW_L, ..., \mW_1)} \|\mW_{i^*:1}\mX_{\vepsilon} - \widehat{\mE}\mX\|_F^2
        \label{derivation: MSE SnT proof identity 8}
    \end{equation}
    
    The equality can only be achieved by solution(s) of the following form:
    \begin{equation}
    \begin{aligned}
        (\mW_L, ..., \mW_1) \in 
        & \{(\mW_L, ..., \mW_1)| (\mW_L, ..., \mW_1) \text{ minimizes } \widehat{\calL}_{\text{base}}(\mW_L, ..., \mW_1)\} \;\cap \\
        &\{(\mW_L, ..., \mW_1)| (\mW_L, ..., \mW_1) \text{ minimizes } \|\mW_{i^*:1}\mX_{\vepsilon} - \widetilde{\mW}_{i^*:1}\mX\|_F^2\}
        \label{derivation: MSE SnT proof identity 7}
    \end{aligned}
    \end{equation}
    
    Eventually we will show that this intersection is nonempty, and the solutions take on a specific form. 
    
    Let's start by examining the second set in the intersection above. For $(\mW_L, ..., \mW_1)$ to belong to the second set, we only have one unique choice for the product of the matrices in the tuple $(\mW_{i^*}, ..., \mW_1)$:
    \begin{equation}
        \widehat{\mW}_{i^*:1} \coloneqq \widetilde{\mW}_{i^*:1}\mX\mX_{\vepsilon}^T(\mX_{\vepsilon}\mX_{\vepsilon}^T)^{-1}
    \end{equation}
    This is just the global minimizer of the loss $\|\mW_{i^*:1}\mX_{\vepsilon} - \widetilde{\mW}_{i^*:1}\mX\|_F^2$ with rank constraint no less than $\text{rank}(\widetilde{\mW}_{i^*:1}\mX\mX_{\vepsilon}^T(\mX_{\vepsilon}\mX_{\vepsilon}^T)^{-1})$. $\mW_{i^*:1}$ can indeed take on this value, since $\text{rank}(\widetilde{\mW}_{i^*:1}\mX\mX_{\vepsilon}^T(\mX_{\vepsilon}\mX_{\vepsilon}^T)^{-1}) \le p$, while $p$ is the maximum rank $\mW_{i^*:1}$ can take on.
    
    We now need to minimize $\widehat{\calL}_{\text{base}}$, assuming that $\mW_{i^*:1} = \widehat{\mW}_{i^*:1}$:
    \begin{equation}
    \begin{aligned}
        & \argmin{(\mW_L, ..., \mW_{i^*+1})} \|\mW_{L:i^*+1}\widehat{\mW}_{i^*:1}\mX_{\vepsilon} - \mY\|_F^2 \\
        = &\argmin{(\mW_L, ..., \mW_{i^*+1})} \|\mW_{L:i^*+1}\widetilde{\mW}_{i^*:1}\mX\mX_{\vepsilon}^T(\mX_{\vepsilon}\mX_{\vepsilon}^T)^{-1}\mX_{\vepsilon} - \mY\|_F^2
    \end{aligned}
    \end{equation}
    We may simplify the above loss as follows:
    \begin{equation}
    \begin{aligned}
        & \|\mW_{L:i^*+1}\widetilde{\mW}_{i^*:1}\mX\mX_{\vepsilon}^T(\mX_{\vepsilon}\mX_{\vepsilon}^T)^{-1}\mX_{\vepsilon} - \mY\|_F^2 \\
        = & \|\mW_{L:i^*+1}\widetilde{\mW}_{i^*:1}\mX\mX_{\vepsilon}^T(\mX_{\vepsilon}\mX_{\vepsilon}^T)^{-1}\mX_{\vepsilon} - \mY\mX_{\vepsilon}^T(\mX_{\vepsilon}\mX_{\vepsilon}^T)^{-1}\mX_{\vepsilon} + \mY\mX_{\vepsilon}^T(\mX_{\vepsilon}\mX_{\vepsilon}^T)^{-1}\mX_{\vepsilon} - \mY\|_F^2 \\
        = & \|\mW_{L:i^*+1}\widetilde{\mW}_{i^*:1}\mX\mX_{\vepsilon}^T(\mX_{\vepsilon}\mX_{\vepsilon}^T)^{-1}\mX_{\vepsilon} - \mY\mX_{\vepsilon}^T(\mX_{\vepsilon}\mX_{\vepsilon}^T)^{-1}\mX_{\vepsilon}\|_F^2 + \|\mY\mX_{\vepsilon}^T(\mX_{\vepsilon}\mX_{\vepsilon}^T)^{-1}\mX_{\vepsilon} - \mY\|_F^2
        \label{derivation: MSE SnT proof identity 3}
    \end{aligned}
    \end{equation}
    The last equality comes from the Pythagorean theorem, the fact that $\mW_{sol} = \mY\mX_{\vepsilon}^T(\mX_{\vepsilon}\mX_{\vepsilon}^T)^{-1}$ is the solution to the MSE problem $\|\mW\mX_{\epsilon} - \mY\|_F^2$, and the following equilibrium identity
    \begin{equation}
        (\mW_{sol}\mX_{\vepsilon} - \mY)\mX_{\vepsilon}^T = \vzero \implies \text{row}(\mW_{sol}\mX_{\vepsilon} - \mY) \perp \text{row}(\mX_{\vepsilon})
    \end{equation}
    Since the second term $\|\mY\mX_{\vepsilon}^T(\mX_{\vepsilon}\mX_{\vepsilon}^T)^{-1}\mX_{\vepsilon} - \mY\|_F^2$ in \eqref{derivation: MSE SnT proof identity 3} is independent of $\mW_j$ for all $j$, we may discard it in the minimization problem. Therefore, we are left to solve
    \begin{equation}
        \argmin{(\mW_L, ..., \mW_{i^*+1})} \|\mW_{L:i^*+1}\widetilde{\mW}_{i^*:1}\mX\mX_{\vepsilon}^T(\mX_{\vepsilon}\mX_{\vepsilon}^T)^{-1}\mX_{\vepsilon} - \mY\mX_{\vepsilon}^T(\mX_{\vepsilon}\mX_{\vepsilon}^T)^{-1}\mX_{\vepsilon}\|_F^2
        \label{derivation: MSE SnT proof identity 4}
    \end{equation}
    If the set of $(\mW_{L}, ..., \mW_{i^*+1})$ that can make the above loss vanish is nonempty, then they are clearly the only set of minimizers of the loss.
    
    But what exactly does making \eqref{derivation: MSE SnT proof identity 4} zero mean? Since $\widetilde{\mW}_{\textbf{L}}\mX = \mY$, the following rearrangement is true:
    \begin{equation}
    \begin{aligned}
        & \mW_{L:i^*+1}\widetilde{\mW}_{i^*:1}\mX\mX_{\vepsilon}^T(\mX_{\vepsilon}\mX_{\vepsilon}^T)^{-1}\mX_{\vepsilon} - \mY\mX_{\vepsilon}^T(\mX_{\vepsilon}\mX_{\vepsilon}^T)^{-1}\mX_{\vepsilon} \\
        = & \mW_{L:i^*+1}\widetilde{\mW}_{i^*:1}\mX\mX_{\vepsilon}^T(\mX_{\vepsilon}\mX_{\vepsilon}^T)^{-1}\mX_{\vepsilon} - \widetilde{\mW}_{L:i^*+1}\widetilde{\mW}_{i^*:1}\mX\mX_{\vepsilon}^T(\mX_{\vepsilon}\mX_{\vepsilon}^T)^{-1}\mX_{\vepsilon} \\
        = & (\mW_{L:i^*+1} - \widetilde{\mW}_{L:i^*+1})\widetilde{\mW}_{i^*:1}\mX\mX_{\vepsilon}^T(\mX_{\vepsilon}\mX_{\vepsilon}^T)^{-1}\mX_{\vepsilon}
    \end{aligned}
    \end{equation}
    But notice that:
    \begin{equation}
    \begin{aligned}
        & (\mW_{L:i^*+1} - \widetilde{\mW}_{L:i^*+1})\widetilde{\mW}_{i^*:1}\mX\mX_{\vepsilon}^T(\mX_{\vepsilon}\mX_{\vepsilon}^T)^{-1}\mX_{\vepsilon} = \vzero \\
        \iff & (\mW_{L:i^*+1} - \widetilde{\mW}_{L:i^*+1})\widetilde{\mW}_{i^*:1}\mX\mX_{\vepsilon}^T(\mX_{\vepsilon}\mX_{\vepsilon}^T)^{-1}\vx_{test} = \vzero, \forall \vx_{test}\in\mathbb{R}^{d_x}
        \label{derivation: MSE SnT proof identity 5}
    \end{aligned}
    \end{equation}
    To see ``$\implies$'', notice that since $\mX_{\vepsilon}$ is of full column rank, for any $\vx_{test} \in \mathbb{R}^{d_x}$, $\vx_{test} = \mX_{\vepsilon}\valpha$ for some $\valpha\in\mathbb{R}^N$. So
    \begin{equation}
    \begin{aligned}
        &(\mW_{L:i^*+1} - \widetilde{\mW}_{L:i^*+1})\widetilde{\mW}_{i^*:1}\mX\mX_{\vepsilon}^T(\mX_{\vepsilon}\mX_{\vepsilon}^T)^{-1}\vx_{test} \\
        = & (\mW_{L:i^*+1} - \widetilde{\mW}_{L:i^*+1})\widetilde{\mW}_{i^*:1}\mX\mX_{\vepsilon}^T(\mX_{\vepsilon}\mX_{\vepsilon}^T)^{-1}\mX_{\vepsilon}\valpha \\
        = & \vzero\valpha \\
        = & \vzero
    \end{aligned}
    \end{equation}
    The ``$\impliedby$'' direction is obvious.
    
    The condition on $\mW_{L:i^*+1}$ in \eqref{derivation: MSE SnT proof identity 5} is clearly equivalent to the following:
    \begin{equation}
        (\mW_{L:i^*+1} - \widetilde{\mW}_{L:i^*+1})\widetilde{\mW}_{i^*:1}\mX\mX_{\vepsilon}^T(\mX_{\vepsilon}\mX_{\vepsilon}^T)^{-1} = \vzero
        \label{derivation: MSE SnT proof identity 10}
    \end{equation}
    
    Therefore, driving \eqref{derivation: MSE SnT proof identity 4} to zero is equivalent to the condition \eqref{derivation: MSE SnT proof identity 10}. Now, what is the set of $(\mW_L, ..., \mW_{i^*+1})$ that satisfies this condition, and more importantly, is this set even nonempty? We shall prove in the next paragraph that this set is indeed nonempty.
    
    By assumption 2. in the theorem statement, $\text{rank}(\mY\mX_{\vepsilon}^T(\mX_{\vepsilon}\mX_{\vepsilon}^T)^{-1}) \le p$, and since $\widetilde{\mW}_{\textbf{L}}\mX = \mY$, 
    \begin{equation}
        \text{rank}(\widetilde{\mW}_{L:i^*+1}\mP_{\widetilde{\mW}_{i^*:1}\mX\mX_{\vepsilon}^T(\mX_{\vepsilon}\mX_{\vepsilon}^T)^{-1}}) \le \text{rank}(\widetilde{\mW}_{L:i^*+1}\widetilde{\mW}_{i^*:1}\mX\mX_{\vepsilon}^T(\mX_{\vepsilon}\mX_{\vepsilon}^T)^{-1}) \le p
    \end{equation}
    must be true. The first inequality needs some justification, which we will discuss below. But assuming that it is true, we now know that there indeed exists a (set of) $\mW_{L:i^*+1}$ such that \eqref{derivation: MSE SnT proof identity 10} is true, in fact, $\mW_{L:i^*+1} = \widetilde{\mW}_{L:i^*+1}\mP_{\widetilde{\mW}_{i^*:1}\mX\mX_{\vepsilon}^T(\mX_{\vepsilon}\mX_{\vepsilon}^T)^{-1}}$ is an example. 
    
    Going back to the first inequality, it holds because for any $\mA, \mB$ for which their product $\mA\mB$ makes sense, $\text{rank}(\mA\mB)\ge\text{rank}(\mA\mP_{\mB})$. To see this, consider the following situations. Case 1: $\mB$ has linearly independent columns. Then $\text{rank}(\mA\mP_{\mB}) = \text{rank}(\mA\mB(\mB^T\mB)^{-1}\mB^T) \le \text{rank}(\mA\mB)$. Case 2: $\mB$ does not have linearly independent columns. Construct $\overline{\mB}$ from $\mB$ by removing the linearly dependent columns of $\mB$. Notice that $\text{rank}(\mA\mB) = \text{rank}(\mA\overline{\mB})$. But
    \begin{equation}
        \text{rank}(\mA\mP_{\mB}) = \text{rank}(\mA\mP_{\overline{\mB}}) = \text{rank}(\mA\overline{\mB}(\overline{\mB}^T\overline{\mB})^{-1}\overline{\mB}^T) \le \text{rank}(\mA\overline{\mB}) = \text{rank}(\mA\mB)
    \end{equation}

    We now arrive at the fact that there does exist (a set of) $(\mW_L, ..., \mW_{i^*+1})$ that satisfies \eqref{derivation: MSE SnT proof identity 10}, therefore, they form the set of minimizers of \eqref{derivation: MSE SnT proof identity 4}. 
    
    But clearly the identity \eqref{derivation: MSE SnT proof identity 10} which characterizes this set of minimizers is equivalent to
    \begin{equation}
        \mW_{L:i^*+1}\widetilde{\mW}_{i^*:1}\mX\mX_{\vepsilon}^T(\mX_{\vepsilon}\mX_{\vepsilon}^T)^{-1} = \widetilde{\mW}_{L:i^*+1}\widetilde{\mW}_{i^*:1}\mX\mX_{\vepsilon}^T(\mX_{\vepsilon}\mX_{\vepsilon}^T)^{-1}
    \end{equation}
    and because $\widehat{\mW}_{i^*:1} = \widetilde{\mW}_{i^*:1}\mX\mX_{\vepsilon}^T(\mX_{\vepsilon}\mX_{\vepsilon}^T)^{-1}$ and $\widetilde{\mW}_{\textbf{L}}\mX = \mY$, the above equality is equivalent to 
    \begin{equation}
        \mW_{L:i^*+1}\widehat{\mW}_{i^*:1} = \mY\mX_{\vepsilon}^T(\mX_{\vepsilon}\mX_{\vepsilon}^T)^{-1}
        \label{derivation: MSE SnT proof identity 6}
    \end{equation}
    We have now arrived at the point to say that, the following set from \eqref{derivation: MSE SnT proof identity 7} is indeed nonempty
    \begin{equation}
    \begin{aligned}
        & \{(\mW_L, ..., \mW_1)| (\mW_L, ..., \mW_1) \text{ minimizes } \widehat{\calL}_{\text{base}}(\mW_L, ..., \mW_1)\} \;\cap \\
        &\{(\mW_L, ..., \mW_1)| (\mW_L, ..., \mW_1) \text{ minimizes } \|\mW_{i^*:1}\mX_{\vepsilon} - \widetilde{\mW}_{i^*:1}\mX\|_F^2\}
    \end{aligned}
    \end{equation}
    and any $(\mW_L, ..., \mW_1)$ belonging to this intersection must satisfy the property
    \begin{equation}
        \mW_{i^*:1} = \widehat{\mW}_{i^*:1}, \text{ and } \mW_{L:i^*+1}\widehat{\mW}_{i^*:1} = \mY\mX_{\vepsilon}^T(\mX_{\vepsilon}\mX_{\vepsilon}^T)^{-1}
    \end{equation}

    Finally, we can conclude that, due to the nonemptiness of the intersection of the two sets from \eqref{derivation: MSE SnT proof identity 7}, the equality in \eqref{derivation: MSE SnT proof identity 8} is indeed achievable, and every solution $(\mW_L^{\text{st}}, ..., \mW_{1}^{\text{st}})$ achieving the equality satisfies
    \begin{equation}
        \mW^{\text{st}}_{\textbf{L}} = \mY\mX_{\vepsilon}^T(\mX_{\vepsilon}\mX_{\vepsilon}^T)^{-1} = \mW^{\text{base}}_{\textbf{L}}
    \end{equation}
    The proof is complete.
    \end{enumerate}
\end{proof}

\begin{corollary}
If $N \ge d_x$, $\widetilde{\mW}_{\textbf{L}}\mX = \mY$, and $p = \min(d_x, d_y)$ (wide networks), then the global minimizers of MSE and student-teacher are identical.
\end{corollary}
\begin{proof}
The inequality $\text{rank}(\mY\mX_{\vepsilon}^T(\mX_{\vepsilon}\mX_{\vepsilon}^T)^{-1/2}) \le \min(d_y, d_x) = p$ must be true, so the application of the above theorem is legal.
\end{proof}

\subsection{Nonlinear-Teacher-Network Results}
\begin{theorem}[Nonlinear teacher, $N < d_x$]
Denote $\mW^{\text{base}}_i(t)$ and $\mW^{\text{st}}_{i}(t)$ as the weights for the student network trained with the the base loss \eqref{eq: MSE loss, deep}, and the student network trained with the student-teacher loss \eqref{eq: SnT loss, deep}, respectively. 

Let the following assumptions hold:
\begin{enumerate}
    \item Gradient flow is the optimizer;
    \item $N_s < d_x$;
    \item $L=2$;
    \item $\{(\vx_i,\vy_i)\}_{i=1}^{N_s}$ and $\{\vepsilon_i\}_{i=1}^{N_s}$ are all sampled independently, and $\vx$ and $\vepsilon$ are continuous random vectors;
    \item There exists some $\delta > 0$ such that $\|\mW_i^{\text{base}}(0)\|_F \le \delta$ and $\|\mW_i^{\text{st}}(0)\|_F \le \delta$ for all $i$;
    \item The teacher network takes the form $\widetilde{\mW}_2 \sigma(\widetilde{\mW}_1\vx)$, with $\sigma(\cdot)$ being a (nonlinear) entry-wise activation function. Furthermore, assume that $\widetilde{\mW}_2 \sigma(\widetilde{\mW}_1\mX) = \mY$, i.e. the teacher network can perfectly solve the clean training problem.
    \item The $\mW^{\text{base}}_i(0)$'s are initialized with the balanced initialization;
    \item Gradient flow successfully converges to a global minimizer for both the MSE- and ST-trained networks;
    \item The weights $\mW_i^{\text{st}}(t)$ remain in a compact set for $t\in[0,\infty)$. In particular, denote $\|\mW_i^{\text{st}}(t)\|_F \le M, t\in[0,\infty)$.
\end{enumerate}

Then the following is true almost surely:
\begin{equation}
    \lim_{t\to\infty} \|\mW_{\textbf{L}}^{\text{base}}(t) - \mW_{\textbf{L}}^{\text{st}}(t)\|_F \le C\delta
\end{equation}
for some $C$ that is bounded as $\delta$ tends to $0$.
\end{theorem}

\begin{proof}
Note that the only difference in assumption between this theorem and the linear-teacher-network theorem is that, we assume the teacher network has nonlinear activation now, and $\widetilde{\mW}_2 \sigma(\widetilde{\mW}_1\mX) = \mY$. Consider the following two points.
\begin{itemize}
    \item Notice that even though the activation function of the teacher is now nonlinear, $\widetilde{\mW}_2 \sigma(\widetilde{\mW}_1\mX)$ still is the product of two matrices, $\widetilde{\mW}_2\in\mathbb{R}^{d_y\times d_1}$ and $\sigma(\widetilde{\mW}_1\mX)\in\mathbb{R}^{d_1\times d_x}$, therefore, $\mY$ has at most rank $p$. It also follows that $\mU_{p}\mU_{p}^T\mY = \mY$. Noting that $\mX_{\epsilon}$ is full-rank almost surely, it is indeed possible to find $(\mW_2, \mW_1)$ such that $\mW_2\mW_1\mX_{\epsilon} = \mY$. In fact, the base-loss solution set is now
    \begin{equation}
        \{\mW_2,\mW_1| \; \mW_2\mW_1 = \mY(\mX_{\vepsilon}^T\mX_{\vepsilon})^{-1}\mX_{\vepsilon}^T + \mR, \; \text{row}(\mR) \subseteq \text{col}(\mX_{\vepsilon})^{\perp}\}
    \end{equation}
    
    Therefore, using exactly the same argument as in the proof for theorem 1 of the paper, we can show that, $\mW_2^{\text{base}}(t)\mW_1^{\text{base}}(t)$ tends to $\mY(\mX_{\epsilon}^T\mX_{\epsilon})^{-1}\mX_{\epsilon}^T + \mW^{\text{base}}(\delta)$ as $t\to\infty$, with $\|\mW^{\text{base}}(\delta)\|_F\in\calO(p^{1/4}\|\mY(\mX_{\vepsilon}^T\mX_{\vepsilon})^{-1}\mX_{\vepsilon}^T\|_F^{1/2})$ when $\delta$ is sufficiently small.

    \item For $\mW_2^{\text{st}}(t)\mW_1^{\text{st}}(t)$, we note that the global minimizers of the student-teacher loss is that set
    \begin{equation}
    \begin{aligned}
        \{\mW_2\in\mathbb{R}^{d_y\times d_1}, \mW_1\in\mathbb{R}^{d_1\times d_x}| & \; \mW_2\mW_1 = \mY(\mX_{\vepsilon}^T\mX_{\vepsilon})^{-1}\mX_{\vepsilon}^T + (\widetilde{\mW}_2 + \mR_2)\mR_1, \\
        & \; \mR_1\in\mathbb{R}^{d_1\times d_x} \land \, \text{row}(\mR_1) \subseteq \text{col}(\mX_{\vepsilon})^{\perp} \land \, \mR_2\in\mathbb{R}^{d_x \times d_1} \land \, \text{row}(\mR_2) \subseteq \text{col}(\sigma(\widetilde{\mW}_1\mX))^{\perp} \}
        \label{eq: SnT solution, nonlinear teacher}
    \end{aligned}
    \end{equation}
    The ``residue matrix'' $(\widetilde{\mW}_2 + \mR_2)\mR_1$ still satisfies the property that $\text{row}((\widetilde{\mW}_2 + \mR_2)\mR_1)\subseteq\text{col}(\mX_{\epsilon}^{\perp})$. Therefore, the gradient-flow argument for $\mW_2^{\text{st}}(t)\mW_1^{\text{st}}(t)$ still holds, so $\mW_2^{\text{st}}(t)\mW_1^{\text{st}}(t)$ must also tend to $\mY(\mX_{\epsilon}^T\mX_{\epsilon})^{-1}\mX_{\epsilon}^T + \mW^{\text{st}}(\delta)$ as $t\to\infty$, with $\|\mW^{\text{st}}(\delta)\|_F \le M\delta$.
\end{itemize}

Combining the above two results finishes our proof.
\end{proof}

\begin{theorem} [Nonlinear teacher, $N \ge d_x$]
Assume the following:
\begin{enumerate}
    \item $N \ge d_x$, and $\mX_{\epsilon}$ is full rank;
    \item $L \ge 2$ (a general deep linear network);
    \item $p \coloneqq \min_{i\in\{0, ..., L\}}d_i \ge  \text{rank}(\mY\mX_{\vepsilon}^T(\mX_{\vepsilon}\mX_{\vepsilon}^T)^{-1})$
    \item Assume that the teacher takes the form $\widetilde{\mW}\mF(\vx)$, with the output dimension of $\mF(\cdot)$ equal to $d_{i^*}$. Also denote $\mF(\mX)\in\mathbb{R}^{d_{i^*}\times N_s}$ as the features the teacher provides to the student, i.e. the student-teacher loss has the form
    \begin{equation}
        \argmin{\mW_L, ..., \mW_1} \left(\|\mW_{\textbf{L}} \mX_{\vepsilon} - \mY \|_F^2 + \lambda \|\mW_{i^*:1}\mX_{\vepsilon} -\mF(\mX)\|_F^2\right)
        \label{eq: ST loss, nonlinear teacher}
    \end{equation}
    Furthermore, assume that the teacher satisfies $\widetilde{\mW}\mF(\mX) = \mY$.
    \item $\min_{i\in\{0, ..., i^*\}}d_i \ge \text{rank}(\mF(\mX)\mX_{\vepsilon}^T(\mX_{\vepsilon}\mX_{\vepsilon}^T)^{-1})$.
\end{enumerate}

Then the global minimizers
\begin{equation}
    \mW_{\textbf{L}}^{\text{base}} = \mW_{\textbf{L}}^{\text{st}} = \mY\mX_{\vepsilon}^T(\mX_{\vepsilon}\mX_{\vepsilon}^T)^{-1}
\end{equation}
\end{theorem}
\begin{proof}
We prove the theorem in two main steps.

\begin{enumerate}
    \item Since $p \coloneqq \min_{i\in\{0, ..., L\}}d_i \ge  \text{rank}(\mY\mX_{\vepsilon}^T(\mX_{\vepsilon}\mX_{\vepsilon}^T)^{-1})$ is still true, the solution for the base loss does not change from before:
    \begin{equation}
        \mW_{\textbf{L}}^{\text{base}} =  \mY\mX_{\vepsilon}^T(\mX_{\vepsilon}\mX_{\vepsilon}^T)^{-1}
    \end{equation}
    \item For the student-teacher loss, we argue in almost the same way as the linear-teacher case. We still prove that
    \begin{equation}
    \begin{aligned}
        \varnothing \neq &\{(\mW_L, ..., \mW_1)| (\mW_L, ..., \mW_1) \text{ minimizes } \widehat{\calL}_{\text{base}}(\mW_L, ..., \mW_1)\} \;\cap \\
        &\{(\mW_L, ..., \mW_1)| (\mW_L, ..., \mW_1) \text{ minimizes } \|\mW_{i^*:1}\mX_{\vepsilon} - \mF(\mX)\|_F^2\}
    \end{aligned}
    \end{equation}
    Like before, we focus on the second set first. To minimize $\|\mW_{i^*:1}\mX_{\vepsilon} - \mF(\mX)\|_F^2$, due to assumption 5., only one solution exists:
    \begin{equation}
        \mW_{i^*:1} = \widehat{\mW}_{i^*:1} \coloneqq \mF(\mX)\mX_{\vepsilon}^T(\mX_{\vepsilon}\mX_{\vepsilon}^T)^{-1}
    \end{equation}
    
    Now, to obtain the solutions in the intersection of the two sets, we assume $\mW_{i^*:1} = \widehat{\mW}_{i^*:1}$ and check what value $\mW_{L:i^*+1}$ can take on. One particular choice is simply $\mW_{L:i^*+1} = \widetilde{\mW}$, in which case we obtain $\mW_{\textbf{L}} = \mY\mX_{\vepsilon}^T(\mX_{\vepsilon}\mX_{\vepsilon}^T)^{-1}$, which indeed minimizes $\widehat{\calL}_{\text{base}}$. It follows that the above intersection is nonempty, so arguing similarly to the linear-teacher case, we may conclude that 
    \begin{equation}
        \mW_{\textbf{L}}^{\text{base}} = \mW_{\textbf{L}}^{\text{st}} = \mY\mX_{\vepsilon}^T(\mX_{\vepsilon}\mX_{\vepsilon}^T)^{-1}
    \end{equation}
    
\end{enumerate}
\end{proof}

\newpage
\section{Proofs for Theorem 3 in Section 5 of the Paper}
In this section, we shall present the proof for theorem 3 of the paper.

\subsection{Notations, Conventions and Assumptions}

Most of the notations and conventions we use are the same as the ones we use for the previous section. We only emphasize the differences here.

Denote $\mX\in\mathbb{R}^{N\times d_x}$ as the clean design matrix, defined by $[\mX]_{i,:} = \vx_i^T$. Similarly, $\mZ\in\mathbb{R}^{N\times d_x}$ is the noise matrix, defined by $[\mZ]_{i,:} = \vepsilon_i^T$. $\mX_{\epsilon} = \mX + \mZ$ is the noisy training input matrix. The target vector is $\vy\in\mathbb{R}^{d_y}$. Recall that the individual target samples are one-dimensional as we are focusing on linear regression in this section.

Given some index set $S\subseteq\{1, ..., n\}$ and vector $\vbeta\in\mathbb{R}^n$, we use $\vbeta_{S}\in\mathbb{R}^{|S|}$ to denote the sub-vector created by extracting the entries in $\vbeta$ with indices contained in $S$, e.g. given $\vbeta = (1, 5, 2, 10)$ and $S=\{2, 4\}$, then $\vbeta_S = (5, 10)$. Similarly, given a matrix $\mM\in\mathbb{R}^{m\times n}$, we denote $\mM_{S}\in\mathbb{R}^{m\times |S|}$ to be the sub-matrix of $\mM$, created by extracting the columns in $\mM$ with indices contained in $S$.

\begin{figure}[!ht]
	\centering
	\includegraphics[width=1.0\linewidth]{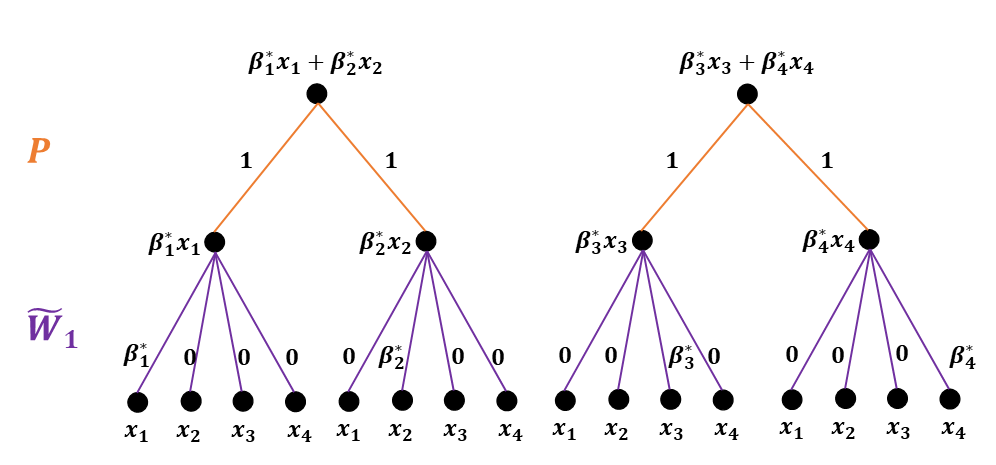}
	\caption{An example of the operation $\mP\widetilde{\mW}_1\vx$. In this example, $d_x=4$, $g=2$. Note that each hidden neuron of the teacher $[\widetilde{\mW}_1]_{i,:}$ only encodes one entry from $\vbeta^*$. The hidden features $[\widetilde{\mW}_1]_{i,:}^T\vx$ are then pooled by the matrix $\mP$. Therefore, in the feature difference loss $\|\mP\mW_1(\vx+\vepsilon)-\mP\widetilde{\mW}_1\vx\|_2^2$, the first group of student neurons sees $\beta^*_1 x_1+\beta^*_2 x_2$, while the second group sees $\beta^*_3 x_3+\beta^*_4 x_4$. Consequently, each group of the student's neurons sees the action of a 2-sparse subset of $\vbeta^*$ on the clean input signal $\vx$.}
	\label{fig: ST lasso illustration}
\end{figure}

We restate the basic assumptions made in the paper, in addition to a few that we specify on the input data:
\begin{enumerate}
    \item The learning problem is linear regression. The ground truth is a linear model $\vbeta^*\in\mathbb{R}^d$, with sparsity level $s$, i.e. only $s$ entries in it are nonzero.
    \item The student and teacher networks are both shallow networks.
    \item We set $m = s$, i.e. the hidden dimension of the networks (i.e. the output dimension of $\mW_1$) is equal to $s$.
    \item We use $\ell_1$ regularization during training.
    \item The student's architecture is $\widetilde{\mW}_2\mP\mW_1(\vx+\vepsilon)$, and teacher's architecture is $\widetilde{\mW}_2\mP\widetilde{\mW}_1\vx$. $\widetilde{\mW}_2\in\mathbb{R}^{1\times m/g}$, and $\mW_{1}, \widetilde{\mW}_1\in\mathbb{R}^{m\times d_x}$. Moreover, $\mP\in\mathbb{R}^{(m/g) \times m}$, $g\in\mathbb{N}$ is a divisor of $m$, and $\mP_{i,j} = 1$ if $j\in\{i g, ..., (i+1) g\}$, and zero everywhere else. Multiplication with $\mP$ essentially \textit{sums every $g$ neurons' output}, similar to how average pooling works in convolutional neural networks.
    \begin{itemize}
        \item In Theorem 3, the weights of the teacher satisfy $[\widetilde{\mW}_2]_i = 1$ for all $i=1, ..., s/g$;  $[\widetilde{\mW}_1]_{i,i} = \beta^*_i$ for $i=1, ..., s$, and the remaining entries are all zeros.
        \item Figure \ref{fig: ST lasso illustration} illustrates $\mP\widetilde{\mW}_1\vx$, for a simple case of $d_x=4$, $g=2$, and the ground truth is $\vbeta^* = (\beta^*_1, ..., \beta^*_4)$. This figure visualizes how the teacher's hidden features are pooled, and presented to the student.
    \end{itemize}
    \item Throughout this whole section, we shall assume that $\vx$ comes from a distribution whose covariance matrix is the identity. The noise $\vepsilon\sim\calN(\vzero, \sigma_{\epsilon}^2\mI_{d_x\times d_x})$, and $\sigma_{\epsilon}<1$.
\end{enumerate}

\subsection{Simplifying the Problem}
We now introduce an equivalent, but more succinct, formulation of the student-teacher learning procedure in Section 5.2, so that we can present the proofs more easily. As we will explain soon, the student-teacher learning setting in section 5.2 of the paper can be decomposed into $s/g$ subproblems of the following form:  for $i\in\{1, ..., s/g\}$,
\begin{equation}
    \vw^{i} = \argmin{\vw}\|\mX_{\epsilon}\vw - \mX\vbeta^{*i}\|_2^2/N_s + \lambda_i \|\vw\|_1
\label{eq: ST problem, LASSO}
\end{equation}
where $\vbeta^{*i}_j = \beta^*_j$ for $j\in\{ig, ..., (i+1)g\}$ and zero everywhere else (it has a sparsity level of $g$). \textbf{We denote the support set $\text{supp}(\vbeta^{*i})$ to be $S_i$} (i.e. it is the set of indices on which $\vbeta^{*i}$ is nonzero).

To see why the above simplified training problem is equivalent to the paper's one, recall that the problem stated in the paper is the following (the feature difference loss itself, without the $\ell_1$ regularization)
\begin{equation}
    \frac{1}{N_s}\sum_{i=1}^{N_s} \Big\|\mP\big[\mW_1(\vx_i+\vepsilon_i) - \widetilde{\mW}_1\vx_i\big]\Big\|_2^2
\end{equation}
But since $\mP_{i,j} = 1$ if $j\in\{i g, ..., (i+1) g\}$, and zero everywhere else, the above loss can be written as a collection of losses independent from each other (enumerated by $i\in\{1,...,s/g\}$):
\begin{equation}
    \frac{1}{N_s}\left\|\mX_{\epsilon}\left(\sum_{j=ig}^{(i+1)g}[\mW_1]_{j,:}^T\right) - \mX\left(\sum_{j=ig}^{(i+1)g} [\widetilde{\mW}_1]_{j,:}^T\right)\right\|_2^2
\end{equation}
For the term $\sum_{j=ig}^{(i+1)g} [\widetilde{\mW}_1]_{j,:}^T$, since in the theorem we assume that $[\widetilde{\mW}_1]_{i,i} = \beta^*_i$ for $i=1, ..., s$, and the remaining entries are all zeros, the vector $\sum_{j=ig}^{(i+1)g} [\widetilde{\mW}_1]_{j,:}^T$'s $ig$-th to $(i+1)g$-th entries are equal to those of $\vbeta^*$ at the same indices, and zero everywhere else. This is where the $\vbeta^{*i}$ came from. 

Let's now add in the $\ell_1$ regularization. For every $i\in\{1, ..., s/g\}$, we have the loss
\begin{equation}
    \frac{1}{N_s}\left\|\mX_{\epsilon}\left(\sum_{j=ig}^{(i+1)g}[\mW_1]_{j,:}^T\right) - \mX\vbeta^{*i}\right\|_2^2 + \lambda_i\left\|\sum_{j=ig}^{(i+1)g}[\mW_1]_{j,:}^T\right\|_1
\end{equation}
Note that we regularize every group of hidden neurons in the student network. One can verify that the minimizer(s) $\sum_{j=ig}^{(i+1)g}[\mW_1]_{j,:}^T$ of the above loss is the same as the minimizer(s) $\vw^i$ of the following loss:
\begin{equation}
    \|\mX_{\epsilon}\vw^i - \mX\vbeta^{*i}\|_2^2/N_s + \lambda_i \|\vw^i\|_1
\label{eq: lasso, simplified test error 1}
\end{equation}

By noting that $\widetilde{\mW}_2$'s entries are all $1$'s, the simplification of the testing loss $\E\left[\left(\widetilde{\mW}_2\mP\mW_1(\vx+\vepsilon) - \vbeta^{*T}\vx\right)^2\right]$ can be argued in a similar way as above, and it simplifies to
\begin{equation}
    \E\left[\left(\sum_{i=1}^{s/g}\vw^{iT}(\vx+\vepsilon) - \vbeta^{*T}\vx\right)^2\right].
\end{equation}

\subsection{Optimal Test Error}
Before going into the theorem and its proof, let us try to understand what the optimal testing error of this regression problem is.

Since we assumed that $\vx$ comes from a distribution with identity covariance, and $\vepsilon\sim\calN(\vzero, \sigma_{\epsilon}^2\mI_{d_x\times d_x})$, the following is true:
\begin{equation}
\begin{aligned}
    \E_{\vx,\vepsilon}[(\vbeta^T(\vx+\vepsilon) - \vbeta^{*T}\vx)^2]
    & = \E_{\vx,\vepsilon}[((\vbeta - \vbeta^{*})^T\vx + \vbeta^T\vepsilon)^2] \\
    & = \E_{\vx,\vepsilon}[((\vbeta - \vbeta^{*})^T\vx)^2] + 2\E_{\vx,\vepsilon}[(\vbeta - \vbeta^{*})^T\vx)(\vbeta^T\vepsilon)] + \E_{\vx,\vepsilon}[(\vbeta^T\vepsilon)^2] \\
    & = \|\vbeta - \vbeta^{*}\|^2_2 + \sigma_{\epsilon}^2\|\vbeta\|_2^2.
    \label{eq: LASSO, simplified test error 2}
\end{aligned}
\end{equation}
It is then easy to show that the optimal linear model that minimizes the above testing error is as follows:
\begin{equation}
    \vbeta^*_{noise} = \frac{1}{1+\sigma_{\epsilon}^2}\vbeta^*
\end{equation}
Furthermore, the optimal testing error is:
\begin{equation}
\begin{aligned}
    \|\vbeta^*_{noise} - \vbeta^{*}\|^2_2 + \sigma_{\epsilon}^2\|\vbeta^*_{noise}\|_2^2 
     & = \frac{\sigma_{\epsilon}^4}{(1+\sigma_{\epsilon}^2)^2}\|\vbeta^*\|_2^2 + \frac{\sigma_{\epsilon}^2}{(1+\sigma_{\epsilon}^2)^2}\|\vbeta^*\|_2^2 \\
     & = \frac{\sigma_{\epsilon}^2\|\vbeta^*\|_2^2}{1+\sigma_{\epsilon}^2}
\label{eq: optimal test error of noisy input problem}
\end{aligned}
\end{equation}

\subsection{Theorem 3 and Its Proof}
\begin{theorem} [Theorem 3 from paper, detailed version]
\label{thm: ST Lasso, mutual incoherence}
Let the following assumptions hold:
\begin{enumerate}
    \item Assumptions in subsection 3.1 hold;
    \item The number of samples satisfies $N_s \in \Omega \left(g^2 \log(d_x)\right)$. 
    \item $\mX$ is fixed, randomness only comes from the noise $\mZ$.
    \item The columns of $\mX$ satisfy $N_s^{-1}\|\mX_i\|_2^2 \le K_x$ for all $i$, with $K_x \in \calO(1)$.
    \item Let $\mX$ satisfy the property that, with high probability (over the randomness of $\mZ$), $\mX_{\epsilon}$ has the mutual incoherence condition for some $\gamma \in (0, 1)$: for every $i\in\{1,...,s/g\}$, for all $j\notin S_i$,
    \begin{equation}
        \max_{j\notin S_i}\|\mX_{\epsilon,j}^T\mX_{\epsilon, S_i}(\mX_{\epsilon, S_i}^T\mX_{\epsilon, S_i})^{-1}\|_1 \le (1-\gamma)
    \end{equation}
    \item With high probability, for every $S_i$, $\mX_{\epsilon,S_i}^T\mX_{\epsilon,S_i}/N_s$ is invertible, and denote its minimum eigenvalue as $\Lambda_{min}^i$. Furthermore, define $\Lambda_{min} = \min_{i\in\{1, ..., s/g\}} \Lambda_{min}^i$.

\end{enumerate}
Then there exists a choice of $\lambda_i$ for each problem $i$ in \eqref{eq: ST problem, LASSO}, such that with high probability, the overall test error satisfies 
\begin{align}
    & \E\left[\left(\widetilde{\mW}_2\mP\mW_1(\vx+\vepsilon) -\vbeta^{*T}\vx \right)^2\right]  \in 
    \calO\bigg(\frac{1}{\gamma^2\Lambda_{min}^2}\frac{\sigma_{\epsilon}^2\|\vbeta^*\|_2^2}{1+\sigma_{\epsilon}^2} \bigg)
\end{align}
\end{theorem}

\begin{remark} Let us interpret the result and assumptions.
\begin{enumerate}
    \item This theorem's conclusion is slightly stronger than the one presented in the main paper, as we eliminated the factor $g$ in the testing error expression.
    \item Condition 4. can be easily satisfied by many types of random matrices, e.g. it would be satisfied with high probability if $\mX$'s entries are sampled independently from the standard Gaussian distribution.
    \item The invertibility condition is almost trivially true, since if we fix $\mX$ and only allow randomness in $\mZ$, then the columns $\mX_{\vepsilon}$ must be independent from each other, and are continuous random vectors. Therefore, $\mX_{\epsilon,S_i}$ will be full-rank almost surely.
    \item The mutual incoherence condition is a common assumption used in the LASSO literature to ensure basis recovery (ours is modified from the standard one, since unlike the traditional case, we have noise in the input). The types of matrices that satisfy mutual incoherence is discussed in \cite{cai_2011} (section 2 and 4) and \cite{tropp_2006} (see proposition 24). For instance, if $\mX$'s entries were sampled independently from the standard Gaussian, then with $N_s \in \Omega(g^2\log(d_x))$, $\mX_{\vepsilon}$ must satisfy mutual incoherence with high probability in high dimensions (over the randomness of $\mX$ and $\mZ$). Note that there are some subtleties with general iid random matrices that have finite exponential moments, as $\log(d_x)\le o(N_s^{c})$ for some $c >0$ could be needed. The general treatment on this condition is beyond the scope of our work.
    \item Note that the sample complexity $N_s\in\Omega(g^2\log(d_x))$ indicates the ``bare minimum'' to ensure reasonable performance of student-teacher learning. As mentioned in the previous point, \textit{more samples are always better}, e.g. if we instead pick $g^2\log(d_x)\le o(N_s^{c})$ for some small $c>0$, then we could get better testing error in the end.
\end{enumerate}
\end{remark}
\begin{proof}
The proof of this theorem follows directly from lemma \ref{lemma: mutual incoherence, subtask error}. During testing, by equations \eqref{eq: lasso, simplified test error 1} and \eqref{eq: LASSO, simplified test error 2}, we just need to compute
\begin{equation}
     \left\|\sum_{i=1}^{s/g}\vw^i - \vbeta^{*}\right\|^2_2 + \sigma_{\epsilon}^2\left\|\sum_{i=1}^{s/g}\vw^i\right\|_2^2
\end{equation}
By lemma \ref{lemma: mutual incoherence, subtask error}, with high probability, for all $i\in\{1,...,s/g\}$, $\text{supp}(\vw^i)\subseteq S_i$, therefore, the above loss can be written as
\begin{equation}
    \left\|\sum_{i=1}^{s/g}(\vw^i - \vbeta^{*i})\right\|^2_2 + \sigma_{\epsilon}^2\left\|\sum_{i=1}^{s/g}\vw^i\right\|_2^2
\end{equation}
Furthermore, since $S_i\cap S_j = \varnothing$ for every $i\neq j$, the above can be written as
\begin{equation}
   \sum_{i=1}^{s/g}\left( \left\|\vw^i - \vbeta^{*i}\right\|^2_2 + \sigma_{\epsilon}^2\left\|\vw^i\right\|_2^2\right)
\end{equation}
Again, by lemma \ref{lemma: mutual incoherence, subtask error}, the above can be bounded with
\begin{equation}
   \sum_{i=1}^{s/g}\left( \left\|\vw^i - \vbeta^{*i}\right\|^2_2 + \sigma_{\epsilon}^2\left\|\vw^i\right\|_2^2\right) \in \calO\bigg(\frac{1}{\gamma^2\Lambda_{min}^2}\frac{\sigma_{\epsilon}^2\|\vbeta^*\|_2^2}{1+\sigma_{\epsilon}^2} \bigg)
\end{equation}

\end{proof}

\subsection{Main Lemmas}
\textbf{We denote $H_{\text{thm}}$ as the intersection of the high-probability events (over the randomness of $\mZ$) from the theorem's assumptions.} In other words, $H_{thm}$ contains the events described in assumptions 5. and 6. in the theorem's statement.

Now, according to the assumptions of Theorem 3, $H_{\text{thm}}$ is assumed to happen with high probability. In the following, we will show that with high probability the solutions $\{\vw^i\}_{i=1}^{s/g}$ will also exhibit certain desirable properties. We will establish these results by showing that the intersection of $H_{\text{thm}}$ and the event that these properties hold has a probability very close to $\mathbb{P}(H_{\text{thm}})$. 

\begin{lemma}
Let the assumptions in the theorem hold. For every $i\in\{1, ..., s/g\}$, choose the $\lambda_i$ in problem \eqref{eq: ST problem, LASSO} as follows
\begin{equation}
    \lambda_i = \frac{20}{\gamma}\sqrt{\frac{\log(d_x)\sigma_{\epsilon}^2\|\vbeta^{*i}\|_2^2 K_x}{N_s}}.
\end{equation}
Then with probability at least $\mathbb{P}(H_{\text{thm}})-5\exp(-c\log(d_x))$ with $c \ge 1$, the following are both true: (i) event $H_{\text{thm}}$ happens; (ii) for all $i,$ the solution $\vw^i$ is unique, and $\text{supp}(\vw^i)\subseteq S_i$ is true. 
\end{lemma}
\begin{proof}

We adopt the approach of the primal dual witness method in \cite{wainwright_2009}. Notice that our optimization problem \eqref{eq: ST problem, LASSO} can be rewritten as (since $\mX_{\epsilon} = \mX+\mZ$)
\begin{equation}
    \|\mX_{\epsilon}(\vw - \vbeta^{*i}) + \mZ\vbeta^{*i}\|_2^2 + \lambda_i\|\vw\|_1.
\end{equation}
It has the same form as theirs (the only difference is that the noise term for us is $-\mZ\vbeta^{*i}$ , while for them it is a noise vector that is independent from the design matrix). Hence, we may directly apply lemmas 2(a) and 3(a) in \cite{wainwright_2009}. Therefore, it suffices for us to prove that, with high probability, for every $i\in\{1, ..., s/g\}$,  $(\mX_{\epsilon, S_i}^T\mX_{\epsilon, S_i})^{-1}$ exists, and the following is true:
\begin{equation}
    \max_{j\notin S_i}\left|\mX_{\epsilon, j}^T\left[\mX_{\epsilon, S_i}(\mX_{\epsilon, S_i}^T\mX_{\epsilon, S_i})^{-1}\vh_{S_i} + \mP_{\mX_{\epsilon, S_i}}^{\perp}\left(\frac{-\mZ\vbeta^{*i}}{\lambda_i N_s}\right)\right]\right|< 1
\label{eq: basis recov lemma, ineq to prove}
\end{equation}
where $\vh_{S_i}$ is a subgradient vector for the $\ell_1$ norm coming from the primal dual witness construction (\cite{wainwright_2009} equation (10)), so $\|\vh_{S_i}\|_{\infty}\le 1$. Note that $(\mX_{\epsilon, S_i}^T\mX_{\epsilon, S_i})^{-1}$ exists as long as $H_{\text{thm}}$ happens. Additionally, recall that $\mP_{\mX_{\epsilon, S_i}}^{\perp}$ denotes the projection onto the orthogonal complement of the column space of $\mX_{\epsilon, S_i}$.

Apply the triangle inequality to the term on the left of the above inequality. We obtain the upper bound
\begin{equation}
    \text{LHS of \eqref{eq: basis recov lemma, ineq to prove}} \le \max_{j\notin S_i}|\mX_{\epsilon, j}^T\mX_{\epsilon, S_i}(\mX_{\epsilon, S_i}^T\mX_{\epsilon, S_i})^{-1}\vh_{S_i}| + \max_{j\notin S_i}\left|\mX_{\epsilon, j}^T\mP_{\mX_{\epsilon, S_i}}^{\perp}\left(\frac{\mZ\vbeta^{*i}}{\lambda_i N_s}\right)\right|.
\end{equation}

If $H_{\text{thm}}$ happens, we can apply the mutual incoherence condition and H\"older's inequality to obtain:
\begin{equation}
    \max_{j\notin S_i}|\mX_{\epsilon, j}^T\mX_{\epsilon, S_i}(\mX_{\epsilon, S_i}^T\mX_{\epsilon, S_i})^{-1}\vh_{S_i}| \le \max_{j\notin S_i}\|\mX_{\epsilon, j}^T\mX_{\epsilon, S_i}(\mX_{\epsilon, S_i}^T\mX_{\epsilon, S_i})^{-1}\|_1\|\vh_{S_i}\|_{\infty} \le 1- \gamma.
\label{eq: basis recov lemma, consequence of mutual incoherence}
\end{equation}

Then, to show \eqref{eq: basis recov lemma, ineq to prove}, it only remains to show
\begin{equation}
    \max_{j\notin S_i}\left|\mX_{\epsilon, j}^T\mP_{\mX_{\epsilon, S_i}}^{\perp}\left(\frac{\mZ\vbeta^{*i}}{\lambda_i N_s}\right)\right| \le \frac{\gamma}{2}
\end{equation}
holds for all $i$ with probability at least $1-5\exp(-c\log(d_x))$.

First note that 
\begin{equation}
    \left|\mX_{\epsilon, j}^T\mP_{\mX_{\epsilon, S_i}}^{\perp}\left(\frac{\mZ\vbeta^{*i}}{\lambda_i N_s}\right)\right| \le \left|\mX_j^T\mP_{\mX_{\epsilon, S_i}}^{\perp}\left(\frac{\mZ\vbeta^{*i}}{\lambda_i N_s}\right)\right| + \left|\mZ_j^T\mP_{\mX_{\epsilon, S_i}}^{\perp}\left(\frac{\mZ\vbeta^{*i}}{\lambda_i N_s}\right)\right|.
\label{eq: noise term}
\end{equation}
The first term on the right hand side (inside the absolute value) is zero-mean sub-Gaussian with parameter at most (by lemma \ref{lemma: subgaussian basic property})
\begin{equation}
    (1/\lambda_i^2N_s^2)\sigma_{\epsilon}^2\|\vbeta^{*i}\|_2^2\|\mP_{\mX_{\epsilon, S_i}}^{\perp}\mX_j\|_2^2 \le (1/\lambda_i^2N_s)\sigma_{\epsilon}^2\|\vbeta^{*i}\|_2^2 K_x
\end{equation}
where we recall from the theorem's assumption that, the columns of $\mX$ satisfy $N_s^{-1}\|\mX_i\|_2^2 \le K_x$ for all $i$, with $K_x \in \calO(1)$. We also made use of the fact that the spectral norm of projection matrices is 1.

Therefore, the following is true in general:
\begin{equation}
    \mathbb{P}\left(\left|\mX_{j}^T\mP_{\mX_{\epsilon, S_i}}^{\perp}\left(\frac{\mZ\vbeta^{*i}}{\lambda_i N_s}\right)\right| > \gamma/4\right) \le 2\exp\left(-\frac{\lambda_i^2 N_s}{\sigma_{\epsilon}^2\|\vbeta^{*i}\|_2^2 K_x} \frac{\gamma^2}{32}\right).
\end{equation}

To ensure the inequality over all $j \notin S_i$, we apply the union bound and obtain: 
\begin{equation}
    \mathbb{P}\left(\max_{j\notin S_i}\left|\mX_{j}^T\mP_{\mX_{\epsilon, S_i}}^{\perp}\left(\frac{\mZ\vbeta^{*i}}{\lambda_i N_s}\right)\right| > \gamma/4\right) \le 2\exp\left(-\frac{\lambda_i^2 N_s}{\sigma_{\epsilon}^2\|\vbeta^{*i}\|_2^2 K_x}\frac{\gamma^2}{32} + \log(d_x - g)\right).
\end{equation}
Our choice of $\lambda_i$ ensures that the above probability is upper bounded by $2\exp(-12\log(d_x))$.

Now we deal with the second term on the right-hand side of \eqref{eq: noise term}:
\begin{equation}
    \left|\mZ_j^T\mP_{\mX_{\epsilon, S_i}}^{\perp}\left(\frac{\mZ\vbeta^{*i}}{\lambda_i N_s}\right)\right| = \left|\mZ_j^T\mP_{\mX_{\epsilon, S_i}}^{\perp}\left(\frac{\sum_{k\in S_i}\mZ_k \beta^*_k}{\lambda_i N_s}\right)\right|.
\end{equation}
Note that $\mZ_j$ for $j\notin S_i$ is independent from $\mP_{\mX_{\epsilon, S_i}}^{\perp}\left(\frac{\sum_{k\in S_i}\mZ_k \beta^*_k}{\lambda_i N_s}\right)$ (the only random terms in it are the $\mZ_k$'s with $k\in S_i$). Therefore, this second term also is zero mean, and in fact has a Gaussian-type tail bound. In particular, denoting $\vv = \mP_{\mX_{\epsilon, S_i}}^{\perp}\left(\frac{\sum_{k\in S_i}\mZ_k \beta^*_k}{\sqrt{N_s}}\right)$, we can write
\begin{equation}
    \mZ_j^T\mP_{\mX_{\epsilon, S_i}}^{\perp}\left(\frac{\mZ\vbeta^{*i}}{\lambda_i N_s}\right) = \mZ_j^T\vv = \left(\frac{1}{\sqrt{N_s}\lambda_i}\mZ_j^T\frac{\vv}{\|\vv\|_2}\right)\|\vv\|_2.
\end{equation}
Notice that due to the rotational invariance of $\mZ_j$, the inner product now produces a Normal random variable regardless of what $\vv$ is. Furthermore,
\begin{equation}
\begin{aligned}
    \mathbb{P}\left(\left|\mZ_j^T\mP_{\mX_{\epsilon, S_i}}^{\perp}\left(\frac{\sum_{j\in S_i}\mZ_j \beta^*_j}{\lambda_i N_s}\right)\right| > \frac{\gamma}{4}\right) 
    \le & \mathbb{P}\left(\left|\frac{1}{\sqrt{N_s}\lambda_i}\mZ_j^T\frac{\vv}{\|\vv\|_2} \right| > \frac{\gamma}{4}\frac{1}{2\sigma_{\epsilon}\|\vbeta^{*i}\|_2}\right) \\
    & + \mathbb{P}\left(\|\vv\|_2 \ge 2\sigma_{\epsilon}\|\vbeta^{*i}\|_2\right).
\end{aligned}
\end{equation}
Let's bound the first probability. Since $\frac{\mZ_j^T\vv}{\|\vv\|_2\sqrt{N_s}\lambda_i}$ is zero-mean sub-Gaussian with parameter at most $\sigma_{\epsilon}^2/(\lambda_i^2 N_s)$, by lemma \ref{lemma: subgaussian basic property} and union bound we have
\begin{equation}
    \mathbb{P}\left(\max_{j\notin S_i}\left|\frac{1}{\sqrt{N_s}\lambda_i}\mZ_j^T\frac{\vv}{\|\vv\|_2} \right| > \frac{\gamma}{4}\frac{1}{2\sigma_{\epsilon}\|\vbeta^{*i}\|_2}\right) \le 2\exp\left(-\frac{\gamma^2}{128}\frac{N_s\lambda_i^2}{\sigma_{\epsilon}^4\|\vbeta^{*i}\|_2^2} + \log(d_x-g)\right).
\end{equation}
With our choice of $\lambda_i$, we can upper bound the above probability by $2\exp(-2\log(d_x))$.

The second probability can be bounded with
\begin{equation}
    \mathbb{P}\left(\|\vv\|_2 \ge 2\sigma_{\epsilon}\|\vbeta^{*i}\|_2\right) \le \mathbb{P}\left(\|\mZ\vbeta^*\|_2^2/N_s \ge 2\sigma_{\epsilon}^2\|\vbeta^{*i}\|_2^2\right)
\end{equation}
since $\left\|\mP_{\mX_{\epsilon, S_i}}^{\perp}\left(\frac{\mZ \vbeta^{*i}}{\sqrt{N_s}}\right)\right\|_2 \le \|\mZ\vbeta^{*i}\|_2/\sqrt{N_s}$. The upper bound on this probability then follows from lemma \ref{lemma: gaussian vector norm concentration, beta*}, and is at most $\exp(-N_s/16)$. With an appropriate choice of $N_s\in\Omega(g^2\log(d_x))$ (sufficiently large constant to multiply with $g^2\log(d_x)$), $\exp(-N_s/16)$ is dominated by $\exp(-2\log(d_x))$.

From the above bounds, we now know that, the following holds with probability at least $1-5\exp(-c'\log(d_x))$ with $c' \ge 2$ (in high dimensions):
\begin{equation}
    \max_{j\notin S_i}\left|\mX_{\epsilon, j}^T\mP_{\mX_{\epsilon, S_i}}^{\perp}\left(\frac{\mZ\vbeta^{*i}}{\lambda_i N_s}\right)\right| \le \gamma/2.
\end{equation}

To ensure that the above inequality holds for all $i$, we take a union bound, and end up with the above inequality holding for all $i$ with probability at least $1-\exp(-c\log(d_x))$ with $c \ge 1$. Combining this with \eqref{eq: basis recov lemma, consequence of mutual incoherence}, with probability at least $\mathbb{P}(H_{\text{thm}})-5\exp(-c\log(d_x))$ with $c \ge 1$, the event $H_{\text{thm}}$ is true, and the following holds for all $i$ (which completes the proof)
\begin{equation}
    \max_{j\notin S_i}\left|\mX_{\epsilon, j}^T\left[\mX_{\epsilon, S_i}(\mX_{\epsilon, S_i}^T\mX_{\epsilon, S_i})^{-1}\vh_{S_i} + \mP_{\mX_{\epsilon, S_i}}^{\perp}\left(\frac{\mZ\vbeta^{*i}}{\lambda_i N_s}\right)\right]\right|< 1 - \frac{\gamma}{2} < 1.
\end{equation}
\end{proof}

\begin{lemma}
\label{lemma: mutual incoherence, subtask error}
Assume the conditions in the theorem hold. Choose
\begin{equation}
    \lambda_i = \frac{20}{\gamma}\sqrt{\frac{\log(d_x)\sigma_{\epsilon}^2\|\vbeta^{*i}\|_2^2 K_x}{N_s}}
\end{equation}
(same as the last lemma). Then, with probability at least $\mathbb{P}(H_{\text{thm}})-5\exp(-c\log(d_x))-3\exp(-\log(d_x))$, the following are both true: (i) $H_{\text{thm}}$ holds; (ii) for all $i\in\{1, ..., s/g\}$, the solution $\vw^i$ is unique and  $\text{supp}(\vw^i)\subseteq\text{supp}(\vbeta^{*i})$ is true, and the following is true:
\begin{equation}
    \E\left[((\vx+\vepsilon)^T\vw^i - \vx^T\vbeta^{*i})^2\right] \le \calO\left(\frac{1}{\gamma^2\Lambda_{min}^2}\frac{\sigma_{\epsilon}^2\|\vbeta^{*i}\|_2^2}{1+\sigma_{\epsilon}^2}\right).
\end{equation}
\end{lemma}
\begin{proof}
Recall from the previous lemma that, with probability at least $\mathbb{P}(H_{\text{thm}})-5\exp(-c\log(d_x))$ (some $c\ge 1$), $H_{\text{thm}}$ holds, \textit{and} for all $i$, $\vw^i$ is unique and $\text{supp}(\vw^i) \subseteq \text{supp}(\vbeta^{*i})$. Let's call this overall event $H_{\text{good}}$.

Assuming $H_{\text{good}}$, the following inequality is true, since $\vw^i$ is the unique solution to the problem \eqref{eq: ST problem, LASSO}:
\begin{align}
    \|\mX_{\epsilon}(\vw^i - \vbeta^{*i}) + \mZ\vbeta^{*i}\|_2^2/N_s + \lambda_i\|\vw^i\|_1 
    & \le \|\mX_{\epsilon}(\vbeta^{*i} - \vbeta^{*i})+ \mZ\vbeta^{*i}\|_2^2/N_s + \lambda_i\|\vbeta^{*i}\|_1 \\
    & = \|\mZ\vbeta^{*i}\|_2^2/N_s + \lambda_i\|\vbeta^{*i}\|_1.
\end{align}
By expanding the first square and cancelling out the $\|\mZ\vbeta^{*i}\|_2^2$, we have:
\begin{equation}
    \|\mX_{\epsilon}(\vw^i - \vbeta^{*i})\|_2^2/N_s + \lambda_i\|\vw^i\|_1  \le \lambda_i\|\vbeta^{*i}\|_1 + 2\vbeta^{*iT}\mZ^T(\mX+\mZ)(\vbeta^{*i} -\vw^i)/N_s.
\end{equation}
Now, note that $\|\vw^i\|_1 = \|\vw^i - \vbeta^{*i} + \vbeta^{*i}\|_1 \ge \|\vbeta^{*i}\|_1 - \|\vw^i - \vbeta^*\|_1$. Therefore, the above inequality leads to
\begin{equation}
    \|\mX_{\epsilon}(\vw^i - \vbeta^{*i})\|_2^2/N_s   \le \lambda_i\|\vw^i-\vbeta^{*i}\|_1+ 2\vbeta^{*iT}\mZ^T(\mX+\mZ)(\vbeta^{*i} -\vw^i)/N_s.
\label{eq: lasso lemma, basic ineq}
\end{equation}

The rest of the proof relies on two main claims. We prove each of them next.

\textbf{Claim 1}: Assuming that $H_{\text{good}}$ happens, with constant $\widehat{C}_1 \in \calO(1)$, for all $i$, 
\begin{equation}
    \lambda_i\|\vw^i-\vbeta^{*i}\|_1 \le \frac{\widehat{C}_1}{\gamma}\sigma_{\epsilon}\|\vbeta^{*i}\|_2\|\vw^i-\vbeta^{*i}\|_2.
\end{equation}

\textbf{Proof of Claim 1}:
Recall that we require $\lambda_i = \frac{20}{\gamma}\sqrt{\frac{\log(d_x)\sigma_{\epsilon}^2\|\vbeta^{*i}\|_2^2 K_x}{N_s}}$. Therefore,
\begin{equation}
    \lambda_i\|\vw^i-\vbeta^{*i}\|_1 = \left(\frac{20}{\gamma}\sqrt{\frac{g\log(d_x) K_x}{N_s}}\right)\sigma_{\epsilon}\|\vbeta^{*i}\|_2\|\vw^i-\vbeta^{*i}\|_2
\end{equation}
where we used the fact that $\|\cdot\|_1 \le \sqrt{g}\|\cdot\|_2$ for $g$-dimensional vectors.

By making an appropriate choice of $N_s\in\Omega(g^2\log(d_x))$ (choosing a sufficiently large constant to multiply with $g^2\log(d_x)$), we have that the following is true:
\begin{equation}
    \lambda_i\|\vw^i-\vbeta^{*i}\|_1 \le \frac{\widehat{C}_1}{\gamma}\sigma_{\epsilon}\|\vbeta^{*i}\|_2\|\vw^i-\vbeta^{*i}\|_2.
\end{equation}
\qed

\textbf{Claim 2}: With probability at least $\mathbb{P}(H_{\text{thm}})-\exp(-c\log(d_x))-3\exp(-\log(d_x))$ and constant $\widehat{C}_2 \in \calO(1)$, $H_{\text{good}}$ holds, and for all $i$,
\begin{equation}
\begin{aligned}
    2\vbeta^{*iT}\mZ^T(\mX+\mZ)(\vbeta^{*i} -\vw^i)/N_s 
    \le \widehat{C}_2 \sigma_{\epsilon}^2\|\vbeta^{*i}\|_2\|\vbeta^{*i} -\vw^i\|_2.
\end{aligned}
\end{equation}

\textbf{Proof of Claim 2}: 
We first note that, if $H_{\text{good}}$ happens, then
:
\begin{equation}
\begin{aligned}
    & 2\vbeta^{*iT}\mZ^T(\mX+\mZ)(\vbeta^{*i} -\vw^i)/N_s \\ 
    = & 2\vbeta_{S_i}^{*iT}\mZ_{S_i}^T(\mX_{S_i}+\mZ_{S_i})(\vbeta^{*i}_{S_i} -\vw^i_{S_i})/N_s \\
    \le & 2|\vbeta_{S_i}^{*iT}\mZ_{S_i}^T\mX_{S_i}(\vbeta^{*i}_{S_i} -\vw^i_{S_i})|/N_s + 2|\vbeta_{S_i}^{*iT}\mZ_{S_i}^T\mZ_{S_i}(\vbeta^{*i}_{S_i} -\vw^i_{S_i})|/N_s.
    \label{eq: central inequality, ST l2 error}
\end{aligned}
\end{equation}

The rest of the proof has two main steps, and the claimed inequality is a direct consequence of the results from the two steps and the above inequality:

\begin{enumerate}
    \item For the first term on the right of the above inequality, consider the following basic upper bound: suppose $\vx_1, \vx_2 \in \mathbb{R}^p$, and $\mM\in\mathbb{R}^{p\times p}$, then
    \begin{equation}
    \begin{aligned}
        |\vx_1^T\mM\vx_2|
        = & |\sum_{i=1}^p\sum_{j=1}^{p} M_{i,j}x_{1,i}x_{2,j}| \\
        \le & \sum_{i=1}^p\sum_{j=1}^p |M_{i,j}x_{1,i}x_{2,j}| \\
        \le & \max_{1\le i',j'\le p}|M_{i',j'}| |\sum_{i=1}^p\sum_{j=1}^p |x_{1,i}||x_{2,j}| \\
        = & \max_{1\le i',j'\le p}|M_{i',j'}|\|\vx_1\|_1\|\vx_2\|_1.
    \end{aligned}
    \end{equation}
    So we have
    \begin{equation}
        |\vbeta_{S_i}^{*iT}\mZ_{S_i}^T\mX_{S_i}(\vbeta^{*i}_{S_i} -\vw^i_{S_i})|/N_s \le \max_{1\le i',j'\le g}\left|[\mZ_{S_i}^T\mX_{S_i}]_{i',j'}\right|\|\vbeta_{S_i}\|_1\|\vbeta^{*i} -\vw^i\|_1/N_s.
    \end{equation}
    But note that, by lemma \ref{lemma: matrix product infinity norm}, in general (not assuming $H_{\text{good}}$), with probability at least $1 - \exp(-2\log(d_x))$,
    \begin{equation}
        \max_{1\le i,j\le g}|[\mZ_{S_i}^T\mX_{S_i}]_{i,j}|/N_s \le \sigma_{\epsilon}\sqrt{\frac{K_x(4\log(d_x) + 4\log(g))}{N_s}}.
    \end{equation}
    Therefore, if $H_{\text{good}}$ \textit{and} the event of lemma \ref{lemma: matrix product infinity norm} happen, we have
    \begin{equation}
    \begin{aligned}
        & \vbeta^{*iT}\mZ^T\mX(\vbeta^{*i} -\vw^i)/N_s \\
        \le & |\vbeta_{S_i}^{*iT}\mZ_{S_i}^T\mX_{S_i}(\vbeta^{*i}_{S_i} -\vw^i_{S_i})|/N_s \\
        \le & \sigma_{\epsilon}\|\vbeta^{*i}\|_1\times \sqrt{\frac{K_x(4\log(d_x) + 4\log(g))}{N_s}}\|\vbeta^{*i} -\vw^i\|_1 \\
        \le & \sigma_{\epsilon}\|\vbeta^{*i}\|_2\times g\sqrt{\frac{K_x(4\log(d_x) + 4\log(g))}{N_s}}\|\vbeta^{*i} -\vw^i\|_2.
    \end{aligned}
    \end{equation}
    In the last inequality, we used the fact that $\|\cdot\|_1 \le \sqrt{g}\|\cdot\|_2$ for $g$-dimensional real vectors. 
    
    With a proper choice of $N_s\in\Omega(g^2\log(d_x))$ (choosing a large enough constant for multiplying with $g^2\log(d_x)$), the above inequality leads to
    \begin{equation}
        2\vbeta^{*iT}\mZ^T\mX(\vbeta^{*i} -\vw^i)/N_s \le C_1 \sigma_{\epsilon}\|\vbeta^{*i}\|_2 \|\vbeta^{*i} -\vw^i\|_2
    \end{equation}
    for constant $C_1 \in \calO(1)$. 
    
    Taking a union bound over all $i$, in general, the event from lemma \ref{lemma: matrix product infinity norm} is true for all $i$ with probability at least $1-\exp(-\log(d_x))$. Therefore, in general, the above inequality is true for all $i$ with probability at least $\mathbb{P}(H_{\text{thm}})-5\exp(-c\log(d_x))-\exp(-\log(d_x))$, since we need $H_{\text{good}}$ and the union of events (over all $i$) of lemma \ref{lemma: matrix product infinity norm} to both hold.
    
    \item Now we upper bound the second inner product term, $|\vbeta_{S_i}^{*iT}\mZ_{S_i}^T\mZ_{S_i}(\vbeta^{*i}_{S_i} -\vw^i_{S_i})|/N_s$. By lemma \ref{lemma: covariance estimation}, denoting $\mE_i = \mZ_{S_i}^T\mZ_{S_i}/N_s - \sigma_{\epsilon}^2\mI_{g\times g}$, we have that in general (not assuming $H_{\text{good}}$), for some universal constant $C_2$, with probability at least $1-2\exp(-2\log(d_x))$,
    \begin{equation}
        \|\mE_i\|_2 \le C_2 \sigma_{\epsilon}\sqrt{
        \frac{g + 2\log(d_x)}{N_s}}
    \end{equation}
    where $\|\cdot\|_2$ represents the spectral norm for square matrices. Note that the above is true \textit{for all} $i$ with probability at least $1-2\exp(-\log(d_x))$. With appropriate choice of $N_s \in \Omega(g^2 \log(d_x))$ (sufficiently large constant to multiply with $g^2 \log(d_x)$), the above expression simplifies to $\|\mE_i\|_2 \le C_2\sigma_{\epsilon}$, for some $C_2 \in \calO(1)$.
    
    Now, if $H_{\text{good}}$ and the union of events (over all $i$) from lemma \ref{lemma: covariance estimation} happen we may write, for all $i$,
    \begin{align}
        |\vbeta_{S_i}^{*iT}\mZ_{S_i}^T\mZ_{S_i}(\vbeta^{*i}_{S_i} -\vw^i_{S_i})|/N_s & = \sigma_{\epsilon}|\vbeta_{S_i}^{*iT}(\vbeta^{*i}_{S_i} -\vw^i_{S_i})| + |\vbeta_{S_i}^{*iT}\mE_i(\vbeta^{*i}_{S_i}  -\vw^i_{S_i})| \\
        & \le  \sigma_{\epsilon}\|\vbeta^{*i}\|_2\|\vbeta^{*i} -\vw^i\|_2 + C_2\sigma_{\epsilon}\|\vbeta^{*i}\|_2\|\vbeta^{*i} -\vw^i\|_2
        \label{eq: LASSO basis recov, second claim ineq}
    \end{align}
    where in the last step we have used $\sigma_{\epsilon}|\vbeta_{S_i}^{*iT}(\vbeta^{*i}_{S_i} -\vw^i_{S_i})| \le \sigma_{\epsilon}\|\vbeta^{*i}\|_2\|\vbeta^{*i} -\vw^i\|_2$ thanks to Cauchy-Schwartz, and $|\vbeta_{S_i}^{*iT}\mE_i(\vbeta^{*i}_{S_i}  -\vw^i_{S_i})| \le C_2\sigma_{\epsilon}\|\vbeta^{*i}\|_2\|\vbeta^{*i} -\vw^i\|_2$ which comes from the following basic inequality. Suppose we have $\vx, \vy \in \mathbb{R}^g$, and $\mM\in\mathbb{R}^{g\times g}$ being symmetric, then $\mM$ has an eigen-decomposition, and we write it as $\mU\mLambda\mU^T$. Then the following holds:
    \begin{equation}
    \begin{aligned}
        \vx^T\mM\vy 
        & = \vx^T\mU\mLambda\mU^T\vy \\
        & = \vx^T\mU\mLambda^{1/2}\mLambda^{1/2}\mU^T\vy \;\;\;\;\;\;\;\; (\text{Define } [\mLambda^{1/2}]_{i,j} = [\mLambda]_{i,j}^{1/2})\\
        & = (\mLambda^{1/2}\mU^T\vx)^T(\mLambda^{1/2}\mU^T\vy) \\
        & \le \|\mLambda^{1/2}\mU^T\vx\|_2 \|\mLambda^{1/2}\mU^T\vy\|_2  \;\;\;\;\;\;\;\; (\text{Cauchy Schwartz}) \\
        & \le \|\mM\|_2\|\mU^T\vx\|_2\|\mU^T\vy\|_2 \\
        & \le \|\mM\|_2\|\vx\|_2\|\vy\|_2  \;\;\;\;\;\;\;\; (\text{Orthogonal matrices preserve $\ell_2$ norm})
    \end{aligned}
    \end{equation}
    
    Using \eqref{eq: LASSO basis recov, second claim ineq}, the inequality below is true with constant $\widetilde{C}_2 \in \calO(1)$:
    \begin{equation}
        2|\vbeta_{S_i}^{*iT}\mZ_{S_i}^T\mZ_{S_i}(\vbeta^{*i}_{S_i} -\vw^i_{S_i})|/N_s \le \widetilde{C}_2\sigma_{\epsilon}\|\vbeta^{*i}\|_2\|\vbeta^{*i} -\vw^i\|_2.
    \end{equation}
    
    Now, let us summarize the probabilities calculated so far. From the previous point we need $H_{\text{good}}$ and the union of events of lemma \ref{lemma: matrix product infinity norm} to be true. Now we also need the union of events of lemma \ref{lemma: covariance estimation} to be true, so we end up with a probability at least $\mathbb{P}(H_{\text{thm}})-5\exp(-c\log(d_x))-\exp(-\log(d_x)) - 2\exp(-\log(d_x))$.
    
    \qed
    
\end{enumerate}

\textbf{Proof of Lemma \ref{lemma: mutual incoherence, subtask error} continued}: With Claim 1 and Claim 2 in hand, we arrive at the fact that, with constant $C \in \calO(1)$, $c\ge 1$ and probability at least $\mathbb{P}(H_{\text{thm}})-5\exp(-c\log(d_x))-3\exp(-\log(d_x))$, $H_{\text{good}}$ holds, and for all $i\in\{1, ..., s/g\}$, the following is true
\begin{equation}
    \|\mX_{\epsilon, S_i}(\vw^i_{S_i} - \vbeta^{*i}_{S_i})\|_2^2/N_s \le \frac{C}{\gamma}\sigma_{\epsilon}\|\vbeta^{*i}\|_2\|\vw^i-\vbeta^{*i}\|_2.
\end{equation}

All that is left is some algebraic manipulations.

Recalling that the minimum eigenvalue of $\mX_{\epsilon, S_i}^T\mX_{\epsilon, S_i}/N_s$ is $\Lambda_{min}^i > 0$, we have the inequality
\begin{equation}
    \|\mX_{\epsilon, S_i}(\vw^i_{S_i} - \vbeta^{*i}_{S_i})\|_2^2/N_s \ge \Lambda_{min}^i\|\vw^i_{S_i} - \vbeta^{*i}_{S_i}\|_2^2 \ge \Lambda_{min}\|\vw^i_{S_i} - \vbeta^{*i}_{S_i}\|_2^2.
\end{equation}

It follows that,
\begin{align}
    & \Lambda_{min}\|\vw^i - \vbeta^{*i}\|_2^2 \le \frac{C}{\gamma}\sigma_{\epsilon}\|\vbeta^{*i}\|_2\|\vw^i-\vbeta^{*i}\|_2  \\
    \implies & \|\vw^i - \vbeta^{*i}\|_2 \le \frac{C}{\gamma\Lambda_{min}}\sigma_{\epsilon}\|\vbeta^{*i}\|_2.
\end{align}
Furthermore, by noting that $\sigma_{\epsilon}\|\vw^i\|_2 \le \sigma_{\epsilon}\|\vw^i - \vbeta^{*i}\|_2 + \sigma_{\epsilon}\|\vbeta^{*i}\|_2$, $1/\sqrt{1+\sigma_{\epsilon}^2}\in\calO(1)$, and using $C$ to absorb $\calO(1)$ constants, we arrive at
\begin{equation}
    \|\vw^i - \vbeta^{*i}\|_2 + \sigma_{\epsilon}\|\vw^i\|_2 \le \frac{C}{\gamma\Lambda_{min}}\frac{\sigma_{\epsilon}}{\sqrt{1+\sigma_{\epsilon}^2}}\|\vbeta^{*i}\|_2.
    \label{eq: LASSO lemma 2, RMSE bound}
\end{equation}

Now, consider the following basic identity:
\begin{equation}
    \sqrt{ \|\vw^i - \vbeta^{*i}\|_2^2 + \sigma_{\epsilon}^2\|\vw^i\|_2^2} \le  \|\vw^i - \vbeta^{*i}\|_2 + \sigma_{\epsilon}\|\vw^i\|_2.
    \label{eq: identity, bound on sum of squares}
\end{equation}

By noting that $\E\left[((\vx+\vepsilon)^T\vw^i - \vx^T\vbeta^{*i})^2\right] = \|\vw^i - \vbeta^{*i}\|_2^2 + \sigma_{\epsilon}^2\|\vw^i\|_2^2$ from section 3.3, and by combining \eqref{eq: LASSO lemma 2, RMSE bound} and \eqref{eq: identity, bound on sum of squares}, we obtain the desired expression in the lemma.

\end{proof}

\subsection{Probability Lemmas}
\begin{lemma}
Let $\mZ$'s entries be sampled from $\calN(0,\sigma_{\epsilon}^2)$ independently, and the columns of $\mX$ satisfy $N_s^{-1}\|\mX_i\|_2^2 \le K_x$ for all $i$. $S_i\subset\{1,...,d_x\}$ is an index set of size $g$. Only $\mZ$ is random, $\mX$ is fixed.

For any $t>0$, with probability at least $1 - \exp(-t^2/2)$, 
\begin{equation}
    \max_{1\le i,j\le g}|[\mZ_{S_i}^T\mX_{S_i}]_{i,j}|/N_s \le \sigma_{\epsilon}\sqrt{\frac{K_x(t^2 + 4\log(g))}{N_s}}.
\end{equation}
\label{lemma: matrix product infinity norm}
\end{lemma}
\begin{proof}
Recall that $[\mZ_{S_i}]_{:,i}\sim\calN(\vzero,\sigma_{\epsilon}^2\mI_{N\times N})$, and since $\mX_{S_i}$ is deterministic, for each $i, j$, $[\mZ_{S_i}^T\mX_{S_i}]_{i,j} = [\mZ_{S_i}]_{:,i}^T[\mX_{S_i}]_{:,j} \sim \calN(0, \sigma_{\epsilon}^2\|[\mX]_{:,j}\|_2^2)$. Therefore, $[\mZ_{S_i}^T\mX_{S_i}]_{i,j}/(\sqrt{N_s K_x}\sigma_{\epsilon})$ is zero-mean sub-Gaussian random variable with its parameter no greater than $1$ for all $i,j$. Now we may apply the union bound and the tail bound for sub-Gaussian random variables (from lemma \ref{lemma: subgaussian basic property}), and arrive at the following result: for any $t>0$, the following holds:
\begin{equation}
\begin{aligned}
    & \mathbb{P}\left(\max_{1\le i,j\le g}|[\mZ_{S_i}^T\mX_{S_i}]_{i,j}|/(\sqrt{N_s K_x}\sigma_{\epsilon}) 
    \ge \sqrt{t^2 + 2\log(g^2)}\right) \\
    \le & 2g^2\exp\left\{-\frac{t^2 + 2\log(g^2)}{2}\right\} \\
    = & 2\exp\left\{-\frac{t^2}{2}\right\}.
\end{aligned}
\end{equation}
But clearly 
\begin{align}
    & \mathbb{P}\left(\max_{1\le i,j\le g}|[\mZ_{S_i}^T\mX_{S_i}]_{i,j}|/(\sqrt{N_s K_x}\sigma_{\epsilon}) 
    \ge \sqrt{t^2 + 2\log(g^2)}\right) \\
    = & \mathbb{P}\left(\max_{1\le i,j\le g}|[\mZ_{S_i}^T\mX_{S_i}]_{i,j}|/\sqrt{N_s} 
    \ge \sigma_{\epsilon}\sqrt{K_x(t^2 + 4\log(g))}\right).
\end{align}
The proof is complete.
\end{proof}

\begin{lemma}
\label{lemma: barvitock lecture note}
Let $\vx\sim\calN(\vzero, \mI_{d\times d})$. Then for any $\delta\ge 0$, the following is true:
\begin{equation}
    \mathbb{P}(\|\vx\|_2^2 \ge d + \delta) \le \left(\frac{d}{d+\delta}\right)^{-d/2}\exp(-\delta /2).
\end{equation}
\end{lemma}
\begin{proof}
This is a relatively standard concentration bound for the $\ell_2$ norm of random Gaussian vectors. We provide its proof for the sake of completeness.

Denote $f_x$ as the probability density function of $\vx$. 

Choose $\lambda = \delta/(d+\delta)$. The following is true:
\begin{equation}
    \|\vx\|_2^2 \ge d + \delta \implies \exp(\lambda\|\vx\|_2^2/2) \ge \exp(\lambda(d + \delta)/2).
\end{equation}
Moreover,
\begin{equation}
    \int_{\mathbb{R}^{d}} \exp(\lambda\|\vx\|_2^2/2)f_x(\vx)d\vx \ge \mathbb{P}(\|\vx\|_2^2 \ge d + \delta)\exp(\lambda(d + \delta)/2).
\end{equation}
Therefore
\begin{equation}
    \mathbb{P}(\|\vx\|_2^2 \ge d + \delta) \le \exp(-\lambda(d + \delta)/2)\int_{\mathbb{R}^{d}} \exp(\lambda\|\vx\|_2^2/2)f_x(\vx)d\vx.
\end{equation}
Explicitly computing the integral on the right-hand-side yields
\begin{equation}
    \mathbb{P}(\|\vx\|_2^2 \ge d + \delta) \le \left(1-\lambda\right)^{-d/2} \exp(-\lambda(d + \delta)/2).
\end{equation}
Substituting $\lambda = \delta/(d+\delta)$ into the expression completes the proof.
\end{proof}

\begin{corollary}
\label{coro: gaussian vector norm concentration}
Let $\vx\sim\calN(\vzero, \mI_{d\times d})$. Then for any $\epsilon\in(0,1)$, the following is true:
\begin{equation}
    \mathbb{P}(\|\vx\|_2^2 \ge (1-\epsilon)^{-1}d ) \le \exp(-\epsilon^2 d/4).
\end{equation}
\end{corollary}
\begin{proof}
In the result of lemma \ref{lemma: barvitock lecture note}, choose $\delta = d\epsilon/(1-\epsilon)$. 

Then $d+\delta = d/(1-\epsilon)$, and we obtain:
\begin{equation}
    \mathbb{P}(\|\vx\|_2^2 \ge d/(1-\epsilon)) \le (1-\epsilon)^{-d/2}\exp\left(-\frac{d}{2}\frac{\epsilon}{1-\epsilon}\right) \le \exp\left(-\frac{d}{2}\left(\frac{\epsilon}{1-\epsilon}+ \log(1-\epsilon)\right)\right).
\end{equation}
We obtain the desired expression by noting that 
\begin{equation}
    \frac{\epsilon}{1-\epsilon}+ \log(1-\epsilon) \ge \epsilon^2/2.
\end{equation}

\end{proof}

\begin{lemma}
\label{lemma: gaussian vector norm concentration, beta*}
Let the entries of $\mW\in\mathbb{R}^{N\times d}$ be independent and have the random distribution $\calN(0, \sigma_{\epsilon}^2)$, and $\vbeta^*\in\mathbb{R}^{d}$.

For any $\delta\in(0, 1)$, with probability at least $1-\exp(-\delta^2 N/4)$, 
\begin{equation}
    \|\mW\vbeta^*\|_2^2/N \le (1-\delta)^{-1}\sigma_{\epsilon}^2\|\vbeta^*\|_2^2.
\end{equation}
\end{lemma}
\begin{proof}
Note that $\mW\vbeta^*/(\sigma_{\epsilon}\|\vbeta^*\|_2) \sim \calN(\vzero, \mI_{N\times N})$. We now invoke the concentration inequality for standard Gaussian random vector from corollary \ref{coro: gaussian vector norm concentration} to obtain
\begin{equation}
    \mathbb{P}\left(\frac{\|\mW\vbeta^*\|_2^2}{N} \ge (1-\delta)^{-1}\sigma_{\epsilon}^2\|\vbeta^*\|_2^2\right)  = \mathbb{P}\left(\frac{\|\mW\vbeta^*\|_2^2}{\sigma_{\epsilon}^2\|\vbeta^*\|_2^2} \ge (1-\delta)^{-1}N\right) \le \exp(-\delta^2N/4).
\end{equation}
\end{proof}

\begin{lemma} [Exercise 4.7.3 in \cite{vershynin_2018}, specialized to iid sub-Guassian vectors]
\label{lemma: covariance estimation}
Let $\vz$ be a zero-mean sub-Gaussian random vector in $\mathbb{R}^g$ with independent and identically distributed entries, with each entry having the same sub-Gaussian random distribution. Moreover, define $\Sigma_Z = \E[\vz\vz^T]$.

Given $\{\vz_i\}_{i=1}^N$, then there exists universal constant $C$ such that, for any $u \ge 0$, the following is true with probability at least $1-2\exp(-u)$:
\begin{equation}
    \left\|\frac{1}{N}\sum_{i=1}^N \vz_i\vz_i^T - \mSigma_Z\right\|_2 \le C\left(\sqrt{\frac{g + u}{N}} + \frac{g + u}{N}\right)\|\mSigma_Z\|_2
\end{equation}
where for matrices, $\|\cdot\|_2$ represents the spectral norm.

\end{lemma}

\begin{lemma}
\label{lemma: subgaussian basic property}
Recall that a zero-mean random variable $X$ is sub-Gaussian if there exists some $\sigma>0$ such that for all $t\in\mathbb{R}$
\begin{equation}
    \E\left[\exp(tX)\right]\le \exp(\sigma^2t^2/2).
\end{equation}

Moreover, $X$ must satisfy (from \cite{wainwright_2009} Appendix A)
\begin{equation}
    \mathbb{P}(|X|>x)\le 2\exp\left(-\frac{x^2}{2\sigma^2}\right).
\end{equation}

An additional useful result is that, if $X_1, ..., X_n$ are independent and zero-mean sub-Gaussian random variables with parameters $\sigma_1^2, ..., \sigma_n^2$, then $\sum_{i=1}^n X_i$ is sub-Gaussian with parameter $\sum_{i=1}^n\sigma_i^2$ (from \cite{buldygin_2000} lemma 1.7).
\end{lemma}

\subsection{Experimental Result}
We carry out the following simple experiment to further support the utility of student-teacher learning over target-based learning. We use the Lasso and LassoLars methods from the scikit-learn library to numerically solve the LASSO problems described below.

In this experiment, we focus on student-teacher learning and target-based LASSO learning. For student-teacher learning, we let $g = 1$. For target-based LASSO, we simply solve the following problem:
\begin{equation}
    \argmin{\vbeta\in\mathbb{R}^{d_x}} \|\mX_{\epsilon}\vbeta - \mX\vbeta^*\|_2^2 + \lambda\|\vbeta\|_1.
\end{equation}

The exact experimental parameters are set as follows. We choose the list $D_x = \{500, 1000, 2000, 4000\}$. For every $d_x \in D_x$, we set the corresponding $\vbeta^*$ with $\beta^*_j = 1.0$ for $j\in\{1, ..., d_x/20\}$, and $0$ everywhere else. So for each $d_x$, $\vbeta^*$ has a sparsity level $s=d_x/20$. The sample size $N_s = 5\log(d_x)$ for every $d_x$. The noise variance $\sigma_{\epsilon}^2 = 0.1$. To solve the base learning problem and the student-teacher learning problem, we run parameter sweep over $\lambda$, and report only the best testing error out of all the $\lambda$'s chosen.

\begin{figure}[!h]
	\centering
	\includegraphics[width=1.0\linewidth]{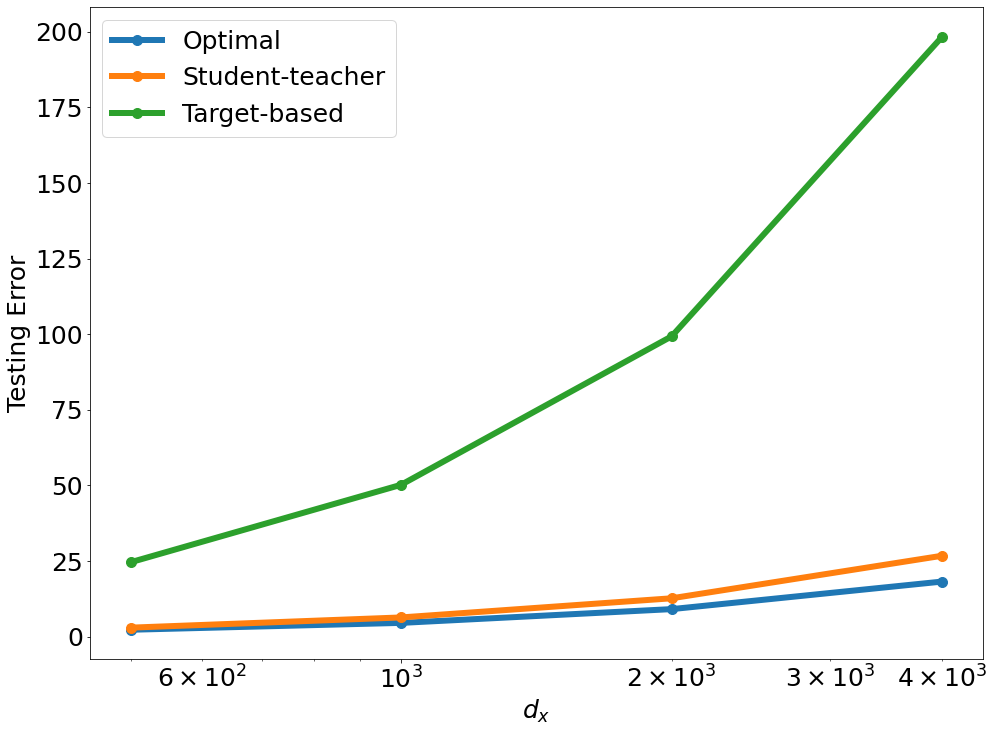}
	\caption{The testing error of the network trained with student-teacher loss and the target-based loss. Optimal testing error is also drawn for comparison. The horizontal axis $d_x$ indicates the data vector dimension, and the vertical axis indicates the testing error of the network. At each $d_x$, we set $s = d_x/20$, $N_s = 5\log(d_x)$. For student-teacher, $g=1$. The noise variance $\sigma_{\epsilon}^2 = 0.1$. We carry out parameter sweep over $\lambda$ for both the target-based and student-teacher problem, and only report the best testing error.}
	\label{fig: LASSO test error}
\end{figure}

Figure \ref{fig: LASSO test error} reports the testing error of the network trained with student-teacher loss and the target-based loss. We also draw the optimal test error curve for comparison. The horizontal axis is $d_x$, and the vertical axis is the testing error. As $d_x$ increases, the testing error of the network trained with target-based LASSO diverges very quickly to infinity, while the testing error of the network trained with the student-teacher loss stays very close to the optimal one.

\section{Second section}

\subfile{main_supplemental}

\end{document}